\documentclass[10pt, doublecolumn, article]{IEEEtran}
	\usepackage[usenames,dvipsnames]{xcolor}
	
	\usepackage{cite}
	\usepackage{amsmath,amssymb,amsfonts,graphicx,bbold,amsthm,mathtools}
	\usepackage{enumitem}
	\usepackage{subfigure,algorithm}
	\usepackage{array}
	\usepackage{xcolor}
	\usepackage{thmtools}
	\declaretheorem{theorem}

	\newtheorem{corollary}{Corollary}[theorem]
	\IEEEoverridecommandlockouts
	\usepackage{times}
	\usepackage{eqparbox}
	\usepackage[noend]{algpseudocode}
	
	\makeatletter
	\def\BState{\State\hskip-\ALG@thistlm}
	\makeatother
	
\begin{document}
	
	\title{Gaussian Processes Over Graphs}
	
	\author{Arun~Venkitaraman, Saikat Chatterjee, Peter H{\"a}ndel\\
	 Department of Information Science and Engineering\\                   
	School of Electrical Engineering and Computer Science \\            
	KTH Royal Institute of Technology,  
	SE-100 44 Stockholm, Sweden                 \\
	arunv@kth.se, sach@kth.se, ph@kth.se
	%\thanks{The authors would like to acknowledge the support received from the Swedish Research Council.}
	}

	\markboth{}
	{}
	\maketitle
	
	\begin{abstract}
		We propose Gaussian processes for signals over graphs (GPG) using the apriori knowledge that the target vectors lie over a graph. We incorporate this information using a graph-Laplacian based regularization which enforces the target vectors to have a specific profile in terms of graph Fourier transform coeffcients, for example lowpass or bandpass graph signals. We discuss how the regularization affects the mean and the variance in the prediction output. In particular, we prove that the predictive variance of the GPG is strictly smaller than the conventional Gaussian process (GP) for any non-trivial graph. We validate our concepts by application to various real-world graph signals. Our experiments show that the performance of the GPG is superior to GP  for small training data sizes and under noisy training. 
	%	Linear regression is one of the most simple yet powerful machine learning approaches.
	%Motivated by the growing interest in analysis of signals that lie over graphs, 

	\end{abstract}
	
	\begin{center}
	%EDICS$-$NEG-ADLE, NEG-APPL, NEG-GRAN, NEG-SPGR.
	\end{center}
	\IEEEpeerreviewmaketitle
	\section{Introduction}
	Gaussian processes are a natural extension of the ubiquitous kernel regression to the Bayesian setting where the regression parameters are modelled as random variables with a Gaussian prior distribution \cite{RasmussenGP}. Given the training observations, Gaussian processes generate posterior probabilities of the target or output for new inputs or observations, as a function of the training data and the input kernel function \cite{Bishop}. Gaussian process models and its variants have been applied in a number of diverse fields such as model predictive control and system analysis \cite{GP_control_1,GP_control_2,GP_control_3, gp_tnn1, gp_tnn8}, latent variable models \cite{GP_latent_1,GP_latent_2, gp_tnn7,gp_tnn10}, multi-task learning \cite{gp_tnn7, multitask_gp0, multitask_gp1}, image analysis and synthesis \cite{GP_image_1,GP_image_2,GP_image_3,GP_image_4}, speech processing \cite{GP_speech_1,GP_speech_2,GP_speech_3}, and magnetic resonance imaging (MRI) \cite{GP_MRI_1,GP_MRI_2}. Gaussian processes have also been extended to a non-stationary regression setting \cite{GP_nonstat_1, gp_tnn2, gp_tnn4} and for regression over complex-valued data\cite{Boloix-TortosaA15}.
	 Recently, Gaussian processes were shown to be useful in training and analysis of deep neural networks, and that a Gaussian process can be viewed as a neural network with a single infinite-dimensional layer of hidden units \cite{deepGP_1,deepGP_2}. The prediction performance of the Gaussian process depends on the availability of training data, progressively improving as the training data size is increased. In many applications, however, we are required to make predictions using a limited number of observations which may be further corrupted with observation noise. For such cases, providing additional structure helps improve the prediction performance of the GP significantly. In this article, we advocate the use of graph signal processing for improving prediction performance in the lack of sufficient and reliable training data. 
	
	We propose Gaussian processes by incorporating the apriori knowledge that the vector-valued target or output vectors lie over an underlying graph.
	% We incorporate graph structure by imposing the target vectors to have a specific profile in terms of its graph Fourier coefficients. 
	This forms a natural Bayesian extension of the kernel regression for graph signals proposed recently in \cite{Arun_kergraph}. 
	%\subsection{Our contributions}
	%We propose Gaussian processes for signals over graphs by imposing the target vectors to lie over a given graph through a graph-Laplacian based regularization.
	 In particular, the target vectors are enforced to follow a pre-specified profile in terms of the graph Fourier coefficients, such being lowpass, bandpass, or high-pass. We show that this in turn translates to a specific structure on the prior distribution of the target vectors. We derive the predictive distribution for a general input given the training observations, and prove that graph signal structure leads to decrease in variance of the predictive distribution. Our hypothesis is that incorporating the graph structure would boost the prediction performance. We validate our hypothesis on various real-world datasets such as temperature measurements, flow-cytometry data, functional MRI data, and tracer diffusion experiment for air pollution studies. Though we consider GPG mainly for undirected graphs characterized by the graph-Laplacian in our analysis, we also discuss how our approach may be extended to handle directed graphs using an appropriate regularization.

	\section{Preliminaries on Graph Signal Processing}
	Graph signal processing or signal processing over graphs deals with extension of several traditional signal processing methods while incorporating the graph structural information \cite{Sandry1,Shuman}. This includes signal analysis concepts such as Fourier transforms\cite{windowedGFT}, filtering \cite{Sandry1, Sandry2}, wavelets \cite{Hammond2011,Wagner2}, filterbanks \cite{vertexfreq,pp_graph1}, multiresolution analysis \cite{Narang2013,ArunSampta15,ArunGHTArxiv}, denoising \cite{deutsch,onuki}, and dictionary learning \cite{Thanou2014, dualgraphelad}, and stationary signal analysis \cite{statgraph1,statgraph5}.
	Spectral clustering and principal component analysis approaches based on graph signal filtering have also been proposed recently \cite{Tremblay2, graphPCA2}. 
	Several approaches have been proposed for learning the graph directly from data \cite{graphlearn1,graphlearn5}. Recently, kernel regression based approaches have also been developed for graph signals \cite{Arun_kergraph,kergraph1,kergraph2}. 
	%We note that to the extent of our knowledge our approach is novel and a Bayesian treatment for prediction for graph signals has not been reported before.
		%\subsection{Graph signal processing}
We now briefly review the relevant concepts from graph signal processing which will be used in our analysis and development. This review has been included to keep our article self-contained.

\subsection{Graph Laplacian and graph Fourier spectrum}
	Consider an undirected graph with $M$ nodes and adjacency matrix $\mathbf{A}$. The $(i,j)$th entry of $\mathbf{A}$ denotes the strength of the edge between the $i$th and $j$th nodes, $\mathbf{A}(i,j)=0$ denoting absence of edge. Since the graph is undirected, we have symmetric edge-weights or $\mathbf{A}=\mathbf{A}^\top$. The graph-Laplacian matrix $\mathbf{L}$ is defined as
	%\begin{equation}
	$\mathbf{L=D-A}$,
	%\nonumber
	%\end{equation}
	where $\mathbf{D}$ is the diagonal degree matrix with $i$th diagonal element given by the sum of the elements in the $i$th row of $\mathbf{A}$\cite{Chung}. $\mathbf{L}$ is symmetric for an undirected graph and by construction has nonnegative eigenvalues with the smallest eigenvalue being equal to zero.
	A graph signal $\mathbf{y}=[y(1)\,y(2)\,\cdots y(M)]^\top\in\mathbb{R}^M$ is an $M$-dimensional vector such that $y(i)$ denotes the value of the signal at the $i$th node, where $\top$ denotes transpose operation. 
The smoothness of $\mathbf{y}$ is measured using
%	\begin{equation}
$l(\mathbf{y})=\displaystyle\frac{	\mathbf{y}^\top\mathbf{L}\mathbf{y}}{\mathbf{y}^\top\mathbf{y}}=\frac{1}{\mathbf{y}^\top\mathbf{y}}{\sum_{\mathbf{A}(i,j)\neq0} \mathbf{A}(i,j)(y(i)-y(j))^2}$.
%\nonumber
%	\end{equation}
	The quantity $l(\mathbf{y})$ is small when $\mathbf{y}$ takes the similar value across all connected nodes, in agreement with what one intuitively expects of a smooth signal. Similarly, $\mathbf{y}$ is a high-frequency or non-smooth signal if it has dissimilar values across connected nodes, or equivalently a large value of $l(\mathbf{y})$.

The eigenvectors of the graph-Laplacian are used to define the notion of frequency of graph signals.
 Let $\{\lambda_i\}_{i=1}^{M}$ and $\{\mathbf{v}_i\}_{i=1}^{M}\in\mathbb{R}^{M}$ denote the eigenvalues and eigenvectors of $\mathbf{L}$, respectively. Then, the eigenvalue decomposition of $\mathbf{L}$ is 
 \begin{eqnarray}
 \mathbf{L}=\mathbf{V}\mathbf{J}_L\mathbf{V}^{\top},
 \label{eq:Form_of_L}
 \end{eqnarray}
 where $\mathbf{J}_L$ is the diagonal eigenvalue matrix of $\mathbf{L}$, such that
 \begin{eqnarray}
 \mathbf{V}&=&[\mathbf{v}_1\,\mathbf{v}_2\cdots \mathbf{v}_M]\,\, \mbox{and}\,\, \mathbf{J}_L=\mbox{diag}(\lambda_1,\lambda_2,\cdots,\lambda_M),\nonumber
 \end{eqnarray}
% The eigenvalues follow the order \cite{Shuman, Chung}:
 %\begin{equation*}
% 0=\lambda_1\leq\lambda_2\leq\cdots\leq\lambda_M
 %\end{equation*}
It is standard practice to order the eigenvectors $\mathbf{v}_i$ of the graph-Laplacian according to their smoothness in terms of $l(\mathbf{v}_i)$. The eigenvectors $\mathbf{v}_i$ are referred to as the graph Fourier transform (GFT) basis vectors since they generalize the notion of the discrete Fourier transform \cite{Shuman}. Let $\lambda_i$ denote the $i$th eigenvalue of $\mathbf{L}$ ordered as 
%\begin{equation*}
$0=\lambda_1\leq\lambda_2\cdots\leq\lambda_M$.
%\end{equation*}
Then, we observe that
%	\begin{equation}
$l(\mathbf{v}_i)=\lambda_i$.
%\nonumber
%\end{equation}
In other words, the eigenvectors corresponding to smaller $\lambda_i$ vary smoothly over the graph and those with large $\lambda_i$ exhibit more variation across the graph. This in turn gives an intuitive frequency ordering of the GFT basis vectors.
The GFT coefficients of a graph signal $\mathbf{x}$ are defined as the inner product of the signal with the GFT basis vectors $\mathbf{V}$, that is, the GFT coefficient vector $\hat{\mathbf{x}}$ is defined~as:
\begin{equation*}
\hat{\mathbf{x}}=\mathbf{V}^\top\mathbf{x}.
\end{equation*}
The inverse GFT is then given by $\mathbf{x}=\mathbf{V}\hat{\mathbf{x}}$.
A graph signal may then be low-pass or smooth, high-pass, band-pass, or band-stop according to the distribution of its GFT coefficients in $\hat{\mathbf{x}}$. 
%\subsection{Filtering}
%The spectral filtering of a graph signal refers to the modification of its GFT coefficients. Let $\mathbf{y}$ denote the output of a filter which modifies the GFT coefficients according to a function $g(\lambda)\geq0$ such that the GFT coeffcients of $\mathbf{y}$ are given by
%\begin{equation*}
%\hat{y}(i)=g(\lambda_i)\hat{{x}}(i).
%\end{equation*}
%Then, we have that
%\begin{equation*}
%\mathbf{y}=\mathbf{V}\mathbf{J}_g\hat{\mathbf{x}}=\mathbf{V}\mathbf{J}_g\mathbf{V}^\top{\mathbf{x}}=\mathbf{G}\mathbf{x},
%\end{equation*}
%where $\mathbf{J}_g$ is the diagonal matrix with entries $g(\lambda_i)$ and  $\mathbf{G}=\mathbf{V}\mathbf{J}_g\mathbf{V}^\top$. Then, using the definition of matrix functions, we have that
%$\mathbf{G}=g(\mathbf{L})$. It is typical to set $g$ to be a polynomial of $\lambda$. Then, we have $\mathbf{G}=g(\mathbf{L})=\sum_{i=1}^K g_i\mathbf{L}^k$ for some $K\leq M$. As $\mathbf{L}$ is symmetric, positive semidefinite and follows the form \eqref{eq:Form_of_L}, it turn out that $\mathbf{G}$ is also symmetric and positive semidefinite. 

	\subsection{Generative model for graph signals}
	We now derive the expressions for the graph signal of a particular GFT profile closest to any given signal, which shall be used in our analysis later.
	For any signal $\mathbf{y}$, a graph signal with a specified graph frequency profile $\mathbf{y}_g$ closest to $\mathbf{y}$ can be obtained by solving the following generative model
\begin{equation*}
\label{GSgen0}
\mathbf{y}_g=\arg\min_{\mathbf{z}}\|\mathbf{y}-\mathbf{z}\|_2^2+\alpha \|\mathbf{J}_p\hat{\mathbf{z}}\|_2^2,\,\,\,\alpha\geq0
\end{equation*}
where $\hat{\mathbf{z}} = \mathbf{V}^\top \mathbf{z}$ is the GFT of $\mathbf{z}$ and $\mathbf{J}_p=\mathrm{diag}(J(1),J(2),\cdots,J(M))$ is the diagonal matrix whose values penalize or promote specific GFT coefficients. For example, if $\mathbf{z}$ is to have energy predominantly in GFT vectors with indices $\{i,j,k,l\}$, we set the corresponding diagonal entries of $\mathbf{J}_p$ to zero and assign large values to the remaining diagonal entries. From the properties of the GFT, we have that
 \begin{eqnarray*}
 \|\mathbf{J}_p\hat{\mathbf{z}}\|_2^2=\hat{\mathbf{z}}^\top\mathbf{J}_p^2\hat{\mathbf{z}} ={\mathbf{z}}^\top\mathbf{V}\mathbf{J}^2_p\mathbf{V}^\top{\mathbf{z}} ={\mathbf{z}}^\top\mathbf{G}{\mathbf{z}},
\end{eqnarray*}
where $\mathbf{G}=\mathbf{V}\mathbf{J}^2_p\mathbf{V}^\top$. 
%In other words, any graph filtering operation may be expressed through a filter of the form $\mathbf{G}=\mathbf{V}\mathbf{J}_G\mathbf{V}^\top$ where $\mathbf{J}_G = \mathbf{J}_p^2$.
Then, the generative model is equivalently expressible as:
\begin{equation}
\label{GSgen1}
\mathbf{y}_g=\arg\min_{\mathbf{z}}\|\mathbf{y}-\mathbf{z}\|_2^2+\alpha {\mathbf{z}}^\top\mathbf{G}{\mathbf{z}},\,\,\,\alpha\geq0.
\end{equation}
Since all the eigenvalues of $\mathbf{G}$ are nonnegative, $\mathbf{G}$ is positive semidefinite giving the unique closed-form solution for $\mathbf{y}_g$ as
\begin{eqnarray}
\mathbf{y}_g=\left(\mathbf{I}_M+\alpha\mathbf{G}\right)^{-1}\mathbf{y}.
\end{eqnarray}
We note that most graph signals encountered in practice are usually smooth over the associated graph. One of the simplest cases of generating a smooth graph signal is to penalize the frequencies linearly by setting ${J}(i)=\sqrt{\lambda_i}$. This in turn corresponds to $\mathbf{G}=\mathbf{V}\mathbf{J}_L\mathbf{T}^\top=\mathbf{L}$ and the smooth graph signal then becomes $\mathbf{y}_g=\left(\mathbf{I}_M+\alpha\mathbf{L}\right)^{-1}\mathbf{y}$.

	\section{Gaussian process over graph}
	 We first develop Bayesian linear regression over graph and then proceed to develop a Gaussian process over graph.
	 
	\subsection{Bayesian linear regression over graph}
	
	We now propose linear regression over graphs. 
	Let $\{\mathbf{x}_n\}_{n=1}^N$ denote the set of $N$ input observations, $n=1,\cdots, N$ and $\mathbf{t}_n\in\mathbb{R}^M$ denote the corresponding vector target values. 
	We model the target as the output of a linear regression such that 
	\begin{equation}
	\label{regprob}
	\mathbf{t}_n=\mathbf{y}(\mathbf{x}_n,\mathbf{W}) + \mathbf{e}_n,
	\end{equation}
	  where 
	  $\mathbf{e}_n$ denotes the additive noise \mbox{vec}tor following a zero mean isotropic Gaussian distribution, that is,
	  \begin{equation}
	  \label{eq:pdf_of_noise}
	  p(\mathbf{e}_n)=\mathcal{N}(\mathbf{e}_n|\mathbf{0},\beta^{-1}\mathbf{I}_M). 
	  \end{equation}
	  Here $\mathbf{I}_M$ denotes the identity matrix of dimension $M$ and $\beta$ is the precision parameter for the noise. The signal $\mathbf{y}$ is
	   % $\mathbf{y}_n(\mathbf{x}_n,\mathbf{W}) $ be the output of the linear regression on function of the input $\phi(\mathbf{x}_n)\in\mathbb{R}^K$ such that
	 \begin{equation}
	 \mathbf{y}(\mathbf{x}_n,\mathbf{W}) =\mathbf{W}^\top\pmb\phi(\mathbf{x}_n) \triangleq \mathbf{y}_n,\nonumber
	 \end{equation}
	 where $\pmb\phi(\mathbf{x})\in\mathbb{R}^K$ is a function of $\mathbf{x}$, $\mathbf{W}\in\mathbb{R}^{K\times M}$ denotes the regression coefficient matrix. 
	 %Throughout our analysis, we specify the noise distributions in terms of the precision parameter which is equal to the inverse of the variance as it simplifies the computations considerably. 
	 We further consider an isotropic Gaussian prior with precision $\gamma$ for the entries of $\mathbf{W}$. Let us use the notation $\tilde{\mathbf{w}} \triangleq \mbox{vec}(\mathbf{W})$ where $\mbox{vec}(\mathbf{W})$ denotes the vectorization of $\mathbf{W}$ obtained by concatenating the columns of $\mathbf{W}$ into a single vector.
Then, we have that
\begin{equation}
\label{eq: w_prior_1}
p(\tilde{\mathbf{w}})=\mathcal{N}(\tilde{\mathbf{w}}|\mathbf{0},\gamma^{-1}\mathbf{I}_{KM}).
\end{equation}
	 
{\color{black} Our principal assumption is that the predicted target vector or regression output $\mathbf{y}_n$ has a graph Fourier spectrum as characterized by the diagonal spectrum matrix $\mathbf{J}_p$. However, 
 $\mathbf{y}_n$ is not necessarily satisfy this requirement for an arbitrary choice of $\mathbf{W}$, since $\pmb\phi(\cdot)$ is also fixed apriori and has not been assumed to be graph-specific. It becomes clear then that the prior on $\tilde{\mathbf{w}}$ should be chosen in a way that it promotes $\mathbf{y}_n$ to have the required graph Fourier spectrum}. We next discuss our strategy for formulating such a prior distribution.
Given the regression output $\mathbf{y}_n$ generated using a fixed $\mathbf{W}$ drawn from (\ref{eq: w_prior_1}), the graph signal $\mathbf{y}_{g,n}$ closest to $\mathbf{y}_n$ is obtained by using the generative model \eqref{GSgen1} with the following result
\begin{eqnarray}
\mathbf{y}_{g,n}&=&\left(\mathbf{I}_M+\alpha\mathbf{G}\right)^{-1}\mathbf{y}_n\nonumber\\
&=&\left(\mathbf{I}_M+\alpha\mathbf{G}\right)^{-1}\mathbf{W}^\top\pmb\phi(\mathbf{x}_n)\nonumber\\
&=&\mathbf{W}_g^\top\pmb\phi(\mathbf{x}_n),\nonumber
\end{eqnarray}
where $\mathbf{W}_g \triangleq \mathbf{W}\left(\mathbf{I}_M+\alpha\mathbf{G}\right)^{-1}$.
% and $\mathbf{I}_M$ denotes the identity matrix of size $M$.
 Let $\mathbf{B}=\left(\mathbf{I}_M+\alpha\mathbf{G}\right)^{-1}$, then we have that \begin{equation}
\label{eq: w_g}
\mathbf{W}_g=\mathbf{W}\mathbf{B}.
\end{equation} 
{\color{black} Thus, we note that regression of $\pmb\phi(\mathbf{x})$ with regression coefficient matrix $\mathbf{W}_g$ produces graph signals possessing the desired graph Fourier profile. In the case of $\mathbf{G}=\mathbf{L}$, this is the same as regression output being smooth over the graph. }
On vectorizing both sides of (\ref{eq: w_g}) and using properties of the Kronecker product \cite{Loan1}, we get that 
 \begin{eqnarray}
\tilde{\mathbf{w}}_g \triangleq \mbox{vec}(\mathbf{W}_g) =\mbox{vec}(\mathbf{W}\mathbf{B})=(\mathbf{B}\otimes\mathbf{I}_K )\tilde{\mathbf{w}},\nonumber
\end{eqnarray}
where $\otimes$ denotes the Kronecker product operation. Note that $\mathbf{B}$ is symmetric as $\mathbf{L}$ and hence, $\mathbf{G}$ is symmetric.
Since $p(\tilde{\mathbf{w}})=\mathcal{N}(\mbox{vec}(\tilde{\mathbf{w}})|\mathbf{0},\gamma^{-1}\mathbf{I}_{KM})$, we get that $\tilde{\mathbf{w}}_g$ is distributed according to

\begin{equation}
p(\tilde{\mathbf{w}}_g)=\mathcal{N}(\tilde{\mathbf{w}_g}|\mathbf{0},\gamma^{-1}(\mathbf{B}^2\otimes\mathbf{I}_K)).
\label{eq: effective_reg_prior}
\end{equation}
This in turn implies that choosing a prior for the regression coefficients of the form \eqref{eq: effective_reg_prior} yields graph signals with specified graph Fourier spectrum over the graph with the Laplacian matrix $\mathbf{L}$. We then pose the regression problem \eqref{regprob} in the following form
	%We model the target as the output of linear regression given by
	\begin{equation}
	\label{regprob}
	\mathbf{t}_n = \mathbf{y}_g(\mathbf{x}_n,\mathbf{W}_g) + \mathbf{e}'_n,\nonumber
	\end{equation}
	We assume that $\mathbf{e}'_n$ follows the same distribution of $\mathbf{e}_n$ in \eqref{regprob}.
Then, the conditional distribution of $\mathbf{t}_n$ is given by 
\begin{equation}
p(\mathbf{t}_n|\mathbf{X},\mathbf{W}_g,\beta)=\mathcal{N}(\mathbf{t}_n|\mathbf{W}_g^\top\pmb\phi(\mathbf{x}_n),\beta^{-1}\mathbf{I}_M).
\end{equation}
We define the matrices $\mathbf{\Phi}$, $\mathbf{Y}_g$, and $\mathbf{T}$ as follows:
 \begin{eqnarray}
 \mathbf{\Phi}&=&[\pmb{\phi}(\mathbf{x}_1)\, \pmb\phi(\mathbf{x}_2)\cdots \pmb\phi(\mathbf{x}_N)]^{\top},\nonumber\\
 \mathbf{Y}_g&=&[\mathbf{y}_{g,1}\, \mathbf{y}_{g,2}\cdots \mathbf{y}_{g,N}]^T,\nonumber\\
 \mathbf{T}&=&[\mathbf{t}_1\, \mathbf{t}_2\cdots \mathbf{t}_N]^T.\nonumber
 \end{eqnarray}
On accumulating all the observations we have
\begin{eqnarray}
\tilde{\mathbf{t}} = \tilde{\mathbf{y}}_g + \tilde{\mathbf{e}}
\label{BLR_eq1}
\end{eqnarray}
where $\tilde{\mathbf{t}} \triangleq \mathrm{vec}(\mathbf{T})$, $\tilde{\mathbf{y}}_g \triangleq \mathrm{vec}(\mathbf{Y}_g)$ and $\tilde{\mathbf{e}}$ is the noise term. Dropping the dependency of the distribution on $\mathbf{X},\mathbf{W}_g,\beta$ for brevity, we have that
\begin{equation}
p(\tilde{\mathbf{t}})=
\mathcal{N}(\tilde{\mathbf{t}}|\tilde{\mathbf{y}},\beta^{-1}\mathbf{I}_{MN}).
\label{BLR_eq2}
\end{equation}
Since $ \mathbf{Y}_g=\mathbf{\Phi}\mathbf{W}_g$, we have that
 \begin{eqnarray}
 \label{BLR_eq3}
 \tilde{\mathbf{y}}&=&\mbox{\mbox{vec}}(\mathbf{\Phi}\mathbf{W}_g)
 = \mbox{\mbox{vec}}(\mathbf{\Phi}\mathbf{W}\mathbf{B})\nonumber\\
 &=&(\mathbf{B}^{\top}\otimes\mathbf{ \Phi})\mbox{vec}(\mathbf{W})=(\mathbf{B}\otimes \mathbf{\Phi})\tilde{\mathbf{w}}.\nonumber
 \end{eqnarray}
 The prior distribution of $\tilde{\mathbf{y}}$ is then
a multivariate Gaussian distribution with mean and covariance given by
\begin{eqnarray}
\label{eq: bayes_lin_reg1}
\mathbb{E}\{\tilde{\mathbf{y}}\}&=&(\mathbf{B}\otimes \mathbf{\Phi})\mathbb{E}\{\tilde{\mathbf{w}}\}=\mathbf{0},\\
\mathbb{E}\{\tilde{\mathbf{y}}\tilde{\mathbf{y}}^\top\}&=&(\mathbf{B}\otimes \mathbf{\Phi})\mathbb{E}\{\tilde{\mathbf{w}}\tilde{\mathbf{w}}^\top\}(\mathbf{B}\otimes \mathbf{\Phi})^\top\nonumber\\
&=&\gamma^{-1}(\mathbf{B}\otimes \mathbf{\Phi})(\mathbf{B}^\top\otimes \mathbf{\Phi}^\top)=\gamma^{-1}(\mathbf{B}^2\otimes \mathbf{\Phi}\mathbf{\Phi}^\top).\nonumber
\end{eqnarray}
We refer to this model as {\it Bayesian linear regression over graph}.

\subsection{Gaussian process over graph}
We next develop Gaussian process over graphs (GPG). We first note the existence of $\mathbf{\Phi}\mathbf{\Phi}^\top$ in \eqref{eq: bayes_lin_reg1} and that the input $\mathbf{x}_n$ enters the equation only in the form of inner products of $\pmb\phi(\cdot)$. The inner product $\pmb\phi(\mathbf{x}_n)^\top\pmb\phi(\mathbf{x}_m)$ is a measure of similarity between the $m$th and $n$th inputs. Keeping this in mind, one may generalize the inner-product $\pmb\phi(\mathbf{x}_m)^\top\pmb\phi(\mathbf{x}_n)$ to any valid kernel function\cite{Bishop} $k(\mathbf{x}_m,\mathbf{x}_n)$ of the inputs $\mathbf{x}_m$ and $\mathbf{x}_n$. The kernel matrix is denoted by $\mathbf{K} = \gamma^{-1} \mathbf{\Phi}\mathbf{\Phi}^\top$, and its $(m,n)$th entry is $k(\mathbf{x}_m,\mathbf{x}_n)$. Following \eqref{BLR_eq1}, \eqref{BLR_eq2}, and \eqref{BLR_eq3} we find that $\tilde{\mathbf{t}}$ is Gaussian distributed with zero mean and following covariance
\begin{equation}
 \mathbf{C}_{g,N}=(\mathbf{B}^2\otimes \mathbf{K})+\beta^{-1}\mathbf{I}_{MN},\label{Cg}
\end{equation}
where we use the first subscript $g$ to denote the graph, and the second subscript $N$ to show explicit dependency on $N$ observations.
%The derivation of the expression for $\mathbf{C}_{g,N}$ with the kernel is similar to that of linear regression in \eqref{eq: bayes_lin_reg2}.
%
%We next compute the predictive distribution of the target vector for the $(N+1)$th input given $N$ observations. 
With $(N+1)$ samples, we have that 	  
	 \begin{eqnarray}
\mathbf{C}_{g,N+1}=\left[\begin{array}{cc}
\mathbf{C}_{g,N} &\mathbf{D}\\
		\mathbf{D}^\top &\mathbf{F}
	\end{array}\right],\nonumber
	 \end{eqnarray}
	 where 
	 \begin{itemize}
	 	\item[] $\mathbf{D}=\mathbf{B}^2\otimes\mathbf{k}$,
	 	\item[] $\mathbf{k}=[k(\mathbf{x}_{N+1},\mathbf{x}_1),k(\mathbf{x}_{N+1},\mathbf{x}_2),\cdots,k(\mathbf{x}_{N+1},\mathbf{x}_N)]^\top$, and
	 	\item[] $\mathbf{F}=k(\mathbf{x}_{N+1},\mathbf{x}_{N+1})\mathbf{B}^2+\beta^{-1}\mathbf{I}_{M}$.
	 \end{itemize}
	 Then, using properties of conditional probability for jointly Gaussian vectors \cite{Bishop}, we have the predictive distribution of GPG as follows
	 \begin{eqnarray}
	 \label{eq: predictive_distribution}
	 p(\mathbf{t}_{N+1}|\mathbf{t}_1,\cdots,\mathbf{t}_N)=\mathcal{N}(\mathbf{t}_{N+1}|\mathbf{\pmb\mu}_{N+1},\mathbf{\Sigma}_{N+1} ),
	 \end{eqnarray}
	  where
	  \begin{itemize}
	  	\item[] $\mathbf{\pmb\mu}_{N+1}=\mathbf{D}^\top\mathbf{C}_{g,N}^{-1}\tilde{\mathbf{t}}$
	  	\mbox{and}  $\mathbf{\Sigma}_{N+1}=\mathbf{F}-\mathbf{D}^\top\mathbf{C}_{g,N}^{-1}\mathbf{D}$.
	  \end{itemize}
%	  On substituting $\mathbf{K}=\gamma^{-1}\mathbf{\Phi\Phi^\top}$ in \eqref{eq: predictive_distribution}, we arrive at the predictive distribution corresponding to linear regression model considered in Section I.
%The mean of the predictive distribution in \eqref{eq: predictive_distribution} is given by
%\begin{equation}
%\mathbf{\pmb\mu}_{N+1}=\mathbf{D}^\top\mathbf{a},
%\end{equation}
%	  where $\mathbf{a}=\displaystyle\mathbf{C}_{g,N}^{-1}\tilde{\mathbf{t}}_{N}\in\mathbb{R}^{MN\times 1}$ and $\mathbf{D}=\mathbf{B}^2\otimes\mathbf{k}\in\mathbb{R}^{M\times MN}$. Let $\bar{\mathbf{b}}_i$ denote the $i$th column of $\mathbf{B}^2$. Then, $i$th column of $\mathbf{D}$ is given by
%	  \begin{equation}
%	  \mathbf{d}_i=\bar{\mathbf{b}}_i\otimes \mathbf{k}.\nonumber
%	  \end{equation}
%	 The $mn$th element of $\mathbf{d}_i$ is given by $\gamma^{-1}\bar{\mathbf{b}}_i(m)k(\mathbf{x}_{N+1},\mathbf{x}_n)$. 
%	   Then, the predictive mean for the $i$th node is given by
%\begin{eqnarray}
%\mathbf{\pmb\mu}_{N+1}(i)&=&\mathbf{d}^T_i\mathbf{a}=\sum_{n=1}^{N}\sum_{m=1}^M \mathbf{d}_i(mn)\mathbf{a}_i({mn})\nonumber\\
%&=&\sum_{n=1}^{N} \sum_{m=1}^M\gamma^{-1} \bar{\mathbf{b}}_i(m)k(\mathbf{x}_{N+1},\mathbf{x}_n)\mathbf{a}_i({mn}).\nonumber
%\end{eqnarray}	  
We note that in the case of completely disconnected graph, which corresponds to the conventional 
Gaussian process (GP), the graph adjacency matrix is equal to the identity matrix and correspondingly, $\mathbf{L=0}$. Then, the covariance of the joint distribution of $\tilde{\mathbf{t}}$ is given by	 
\begin{equation} 
	\mathbf{C}_{c,N}=(\mathbf{I}_M\otimes\mathbf{K})+\beta^{-1}\mathbf{I}_{MN},\nonumber
	\end{equation}
	where the subscript $c$ denotes it being the conventional Gaussian process.
	Correspondingly, the predictive distribution of $\mathbf{t}_{N+1}$ is given by
	 \begin{eqnarray}
	p(\mathbf{t}_{N+1}|\mathbf{t}_1,\cdots,\mathbf{t}_N)=\mathcal{N}(\mathbf{t}_{N+1}|\mathbf{\pmb\mu}_{c,N+1},\mathbf{\Sigma}_{c,N+1} ),\nonumber
	\end{eqnarray}
	where
	\begin{itemize}
		\item[] $\mathbf{\pmb\mu}_{c,N+1}=\mathbf{D}_c^\top\mathbf{C}_{c,N}^{-1}\tilde{\mathbf{t}}$,
		\item[] $\mathbf{\Sigma}_{c,N+1}=\mathbf{F}_c-\mathbf{D}_c^\top\mathbf{C}_{c,N}^{-1}\mathbf{D}_c$,
		\item[]
		$\mathbf{D}_c=\mathbf{I}_M\otimes\mathbf{k}$,
		\item[] $\mathbf{F}_c=(k(\mathbf{x}_{N+1},\mathbf{x}_{N+1})+\beta^{-1})\mathbf{I}_{M}$.
	\end{itemize}
We note that since Bayesian linear regression on graphs developed in Section A is a special case of the Gaussian process on graphs with kernels, we shall refer to the former as {\it Gaussian process-Linear (GPG-L)} and the latter as {\it Gaussian process-Kernel (GPG-K)}. Correspondingly, we refer to the conventional versions as GP-L, and GP-K, respectively.

We next show that use of graph information reduces the variance of the predictive distribution.
	This implies that the GPG models the observed training samples better than conventional GP.
	\begin{theorem}[Reduction in variance]
		Graph structure reduces the variance of the marginal distribution $p(\tilde{\mathbf{t}})$, that is, GPG results in smaller variance of $p(\tilde{\mathbf{t}})$ than GP: 
\begin{eqnarray}
%	MSE_g&<& MSE_r,\,\,\,\mbox{or}\\
\mathrm{tr}\left(\mathbf{C}_{c,N}\right)&>& \mathrm{tr}\left(\mathbf{C}_{g,N}\right).\nonumber
\end{eqnarray}
\begin{proof}
%$\mathbf{ C}^{g,N}\preceq \mathbf{C}^{c,N}$ is the same as having the difference of the matrices $\Delta \mathbf{C}_N:=\mathbf{ C}^{g,N}- \mathbf{C}^{c,N}$ to be negative semidefinite, 
%
%Since trace of a matrix is a linear operation, we are required to prove that  $\mbox{tr}(\Delta \mathbf{C}_N)<0$, where $\Delta \mathbf{C}_N:=\mathbf{ C}^{c,N}- \mathbf{C}^{g,N}$. 
In order to prove the result, we need to show that the trace of $\Delta \mathbf{C}_N:=\mathbf{ C}_{c,N}- \mathbf{C}_{g,N}$ is nonnegative.
Using the properties of trace operation, we have that
\begin{align}
\mathrm{tr}(\Delta\mathbf{C}_N)&=\mathrm{tr}(\mathbf{ C}_{c,N}- \mathbf{C}_{g,N})\nonumber\\
&=\mathrm{tr}((\mathbf{I}_M-\mathbf{B}^2)\otimes\mathbf{K})=\mathrm{tr}(\mathbf{I}_M-\mathbf{B}^2)tr(\mathbf{K})\nonumber
\end{align}
Since $\mathbf{K}$ is positive semidefinite by construction as a kernel matrix, we have that $\mathrm{tr}(\Delta\mathbf{C}_N)\geq0$ if and only if $\mathrm{tr}(\mathbf{I}_M-\mathbf{B}^2)\geq 0$. 
%Let $\{\lambda_i\}_{i=1}^{N}$ and $\{\mathbf{v}_i\}_{i=1}^{N}\in\mathbb{R}^{N}$ denote the eigenvalues and eigenvectors of $\mathbf{L}$, respectively; and
 Let $\{\theta_i\}_{i=1}^{N}$ and $\{\mathbf{u}_i\}_{i=1}^{N}\in\mathbb{R}^{N}$ denote the eigenvalues and eigenvectors  of $\mathbf{K}$, respectively. Then, the eigenvalue decomposition of $\mathbf{G}$ and $\mathbf{K}$ are given by
\begin{eqnarray}
\mathbf{G}=\mathbf{V}\mathbf{J}_G\mathbf{V}^{\top},\,\,\,\,
\mathbf{K}=\mathbf{U}\mathbf{J}_K\mathbf{U}^{\top},\nonumber
\end{eqnarray}
where $\mathbf{J}_K$ and $\mathbf{J}_G$ denote the diagonal eigenvalue matrix of $\mathbf{K}$ and $\mathbf{G}$, respectively, such that
\begin{eqnarray}
\mathbf{V}&=&[\mathbf{v}_1\,\mathbf{v}_2\cdots \mathbf{v}_M]\,\, \nonumber\\
\mathbf{J}_G&=&\mbox{diag}(J^2(1),J^2(2),\cdots,J^2(M)),
\nonumber\\ 
\mathbf{U}&=&[\mathbf{u}_1\,\mathbf{u}_2\cdots \mathbf{u}_N]\,\, \mbox{and}\,\,\nonumber\\
 \mathbf{J}_K&=&\mbox{diag}(\theta_1,\theta_2,\cdots,\theta_N).\qquad \nonumber
\end{eqnarray}
{\color{black}Since $\mathbf{J}_G=\mathbf{J}^2_p=\mathrm{diag}(J^2(1),J^2(2),\cdots J^2(M))$,  we have 
\begin{align}
\mathrm{tr}(\mathbf{I}_M-\mathbf{B}^2)> 0,\nonumber\\
\mathrm{tr}(\mathbf{I}_M-\mathbf{V}\left(\mathbf{I}+\alpha\mathbf{J}_G\right)^{-2}\mathbf{V}^\top)\geq 0,\nonumber\\
\mathrm{tr}(\mathbf{I}_M-(\mathbf{I}_M+\alpha\mathbf{J}_G)^{-2})\geq 0,\nonumber\\
\sum_{i=1}^{M}\left(1-\frac{1}{(1+\alpha J(i))^{2}}\right)\geq 0.\nonumber
\end{align}
Let $s$ denote the graph eigenvalue index corresonding to the  denote the smallest non-zero diagonal value in $\mathbf{J}_p$, the value given by $J(s)$. 
%\footnote{also referred to as the Fiedler value or algebraic connectivity of the graph in spectral graph theory \cite{Chung}}. 
Then, we have that $\displaystyle\frac{1}{(1+\alpha J^2(i))}\leq\frac{1}{(1+\alpha J^2(s))}$ $\forall J^2(i)\neq 0$. Then, a sufficient condition to ensure $\mathrm{tr}(\mathbf{I}_M-\mathbf{B}^2)\geq 0$ is to impose that
\begin{align}
(1+\alpha J^2(s))^{-2}\leq 1,\nonumber\\
%((1-\alpha)+\alpha\lambda_i)^{-1}\leq 1\nonumber\\
(1+\alpha J^2(s))\geq 1,\nonumber\\
\alpha J^2(s)\geq 0.\nonumber
%\alpha\geq 0\nonumber\\
\end{align}
 A graph with atleast one connected subgraph has atleast one eigenvalue of $\mathbf{L}$ strictly greater than 0 \cite{Chung}.  Thus, barring the completely disconnected graph and the pathological case of $\mathbf{J}_p=\mathbf{0}$ (or $\mathbf{G}=\mathbf{0}$), any graph with connections across nodes results in $\alpha J^2(s)>0$ or in other words that the variance {\it strictly} reduces in comparison with the conventional regression, that is,
\begin{equation}
\mathrm{tr}(\Delta\mathbf{C}_N)>0.\nonumber
\end{equation}
In other words, introducing non-zero connections among the nodes ensures that the marginal variance is strictly lesser than that obtained for the conventional regression case. We also note that in order for the reduction in variance to be significant, $\lambda_s$ must be large, which in turn implies that the connected communities within the graph must have strong algebraic connectivity \cite{Chung}. In the case of regression output being smooth graph signals in the sense of $\mathbf{G}=\mathbf{L}$, we have that $J^2(i)=\lambda_i$ and $s=K$ for a graph with $K$-connected components or disjoint subgraphs. This is because the number of zero eigenvalues of $\mathbf{L}$ is equal to the nunber of connected components in the graph \cite{Chung,godsil2001algebraic}.}
\end{proof}
	\end{theorem}
An immediate consequence of the Theorem is the following important Corrollary which informs us that the variance of the predicted target reduces when the graph signal structure is employed:
\begin{corollary}[Reduction in predictive variance]
	GPG-K with a non-trivial graph has strictly smaller predictive variance than GPG-K, that is,
	\begin{equation*}
	\mathrm{tr}\left(\mathbf{\Sigma}_{c,N+1}\right)> \mathrm{tr}\left(\mathbf{\Sigma}_{g,N+1}\right).
	\end{equation*}
	\begin{proof}
		We are required to prove that $\mathrm{tr}\left(\mathbf{\Sigma}_{c,N+1}-\mathbf{\Sigma}_{g,N+1}\right)>0$, or equivalently that $\Delta\pmb\Sigma:=\mathbf{\Sigma}_{c,N+1}-\mathbf{\Sigma}_{g,N+1}$ is a positive semidefinite matrix. We notice that $\Delta \mathbf{C}_{N+1}$ is  given by
		
		\begin{eqnarray}
		\Delta\mathbf{C}_{N+1}=\left[\begin{array}{cc}
		\Delta\mathbf{C}_{N} &\mathbf{D}_c-\mathbf{D}\\
		\mathbf{D}_c^\top-\mathbf{D}^\top &\mathbf{F}_c-\mathbf{F}
		\end{array}\right].\nonumber
		\end{eqnarray}
		We observe that $\Delta\pmb\Sigma_{N+1}$ is then the Schur complement of $\mathbf{F}_c-\mathbf{F}$ in $\Delta\mathbf{C}_{N+1}$. Since the Schur complement of a positive-definite matrix is also positive-definite \cite{Horn}, and we have already proved that $\Delta\mathbf{C}_{N+1}$ is positive-definite from Theorem 1, it follows that 
		\begin{eqnarray}
			\mathrm{tr}\left(\mathbf{\Sigma}_{c,N+1}-\mathbf{\Sigma}_{g,N+1}\right)>0,\,\,\mbox{or}\quad
			\mathrm{tr}\left(\mathbf{\Sigma}_{g,N+1}\right)< \mathrm{tr}\left(\mathbf{\Sigma}_{c,N+1}\right).\nonumber
		\end{eqnarray}
	\end{proof}
	\end{corollary}
By the preceding analysis, we also observe that the variance of the joint distribution $\mathbf{C}_{g,N+1}$, and hence, that of $\pmb\Sigma_{g,N+1}$ is inversely related to the graph regularization  parameter $\alpha$. This is because a large $\alpha$ implies a small value of $\displaystyle\frac{1}{1+\alpha J^2(i)}$ for all $i$, and therefore, a small value of $\mbox{tr}(\mathbf{C}_{g,N+1})$.

\subsection{On the mean vector of predictive distribution}
We now show that the mean of predictive distribution is a graph signal with a graph Fourier spectrum which adheres to the condition imposed by $\mathbf{J}_p$. We demonstrate this by computing the graph Fourier spectrum of the predictive mean.  Using the eigendecompostions of $\mathbf{G}$ and $\mathbf{K}$, we have that
\begin{align*}
\mathbf{\pmb\mu}_{N+1}&=\mathbf{D}^\top\mathbf{C}_{g,N}^{-1}\tilde{\mathbf{t}}\nonumber\\
&=\mathbf{D}^\top(\mathbf{B}^2\otimes \mathbf{K}+\beta^{-1}\mathbf{I}_{MN})^{-1}\tilde{\mathbf{t}}\\
&=\mathbf{D}^\top(\mathbf{V}\left(\mathbf{I}+\alpha\mathbf{J}_G\right)^{-2}\mathbf{V}^\top\otimes \mathbf{U}\mathbf{J}_K\mathbf{U}^\top+\beta^{-1}\mathbf{I}_{MN})^{-1}\tilde{\mathbf{t}}\\
&=\mathbf{D}^\top(\mathbf{V}\otimes\mathbf{U})\mathbf{J}(\mathbf{V}^\top\otimes\mathbf{U}^\top)\tilde{\mathbf{t}}=\mathbf{D}^\top(\mathbf{Z}\mathbf{J}\mathbf{Z}^\top)\tilde{\mathbf{t}}\\
&=\mathbf{D}^\top\sum_{i=1}^{MN}\eta_i\rho_i\mathbf{z}_i\,
\end{align*}
where 
\begin{enumerate}
	\item[] $\mathbf{J}=(\left(\mathbf{I}+\alpha\mathbf{J}_G\right)^{-2}\otimes \mathbf{J}_K+\beta^{-1}\mathbf{I}_{MN})^{-1}$,
	\item[] $\eta_i$ is the $i$th diagonal element of $\mathbf{J}$,
	\item[] $\mathbf{Z}=\mathbf{V}\otimes \mathbf{U}$,
	\item[] $\mathbf{z}_i$ denotes the $i$th column vector of $\mathbf{Z}$, and
	\item[] $\rho_i=\mathbf{z}_i^\top\tilde{\mathbf{t}}$.
\end{enumerate}
Since the $\eta_i$ is a function of some $i1$th eigenvalue of $\mathbf{L}$ and $i2$th eigenvalue of $\mathbf{K}$, we shall alternatively use the notation $\eta_{i1,i2}$ to be more specific in the following analysis. Similarly, $\rho_i$ expressed as $\rho_{i1,i2}$ and the eigenvectors $\mathbf{z}_i=\mathbf{v}_{i1}\otimes\mathbf{u}_{i2}$.
The component of the prediction mean along the graph eigenvector $\mathbf{v}_k$ (or the $k$th graph frequency) is then given by
\begin{align}
\pmb\mu_{N+1}^\top\mathbf{v}_k&=\sum_{i=1}^{MN}\eta_i\rho_i\mathbf{z}^\top_i\mathbf{D}\mathbf{v}_k=\sum_{i=1}^{MN}\eta_i\rho_i\mathbf{z}^\top_i(\mathbf{B}^2\otimes \mathbf{k})\mathbf{v}_k\nonumber\\
&=\sum_{i=1}^{MN}\eta_i\rho_i(\mathbf{v}^\top_{i1}\otimes\mathbf{u}_{i2}^\top)(\mathbf{B}\otimes \mathbf{k})\mathbf{v}_k\nonumber\\
&=\sum_{i=1}^{MN}\eta_i\rho_i(\mathbf{v}^\top_{i1}\mathbf{B}\otimes \mathbf{u}_{i2}^\top\mathbf{k})\mathbf{v}_k\nonumber\\
&=\sum_{i=1}^{MN}\eta_i\rho_i(\mathbf{v}^\top_{i1}\mathbf{V}\left(\mathbf{I}+\alpha\mathbf{J}_G\right)^{-2}\mathbf{V}^\top\otimes \mathbf{u}_{i2}^\top\mathbf{k})\mathbf{v}_k\nonumber\\
&=\sum_{i1=1}^{M}\sum_{i2=1}^N\eta_{i1,i2}\rho_{i1,i2}((1+\alpha J^2(i1))^{-2}\mathbf{v}^\top_{i1}\otimes \mathbf{u}_{i2}^\top\mathbf{k})\mathbf{v}_k\nonumber\\
&=\sum_{i1=1}^{M}\sum_{i2=1}^N\eta_{i1,i2}\rho_{i1,i2}\mathbf{u}_{i2}^\top\mathbf{k}(1+\alpha J^2(i1))^{-2}(\mathbf{v}^\top_{i1})\mathbf{v}_k\nonumber\\
&=\sum_{i2=1}^N\eta_{k,i2}\rho_{k,i2}\mathbf{u}_{i2}^\top\mathbf{k}(1+\alpha J^2(k))^{-2}\mathbf{v}^\top_{k}\mathbf{v}_k\nonumber\\
&=\frac{1}{(1+\alpha J^2(k))^2}\sum_{i=1}^{MN}\eta_{k,i2}\rho_{k,i2}\mathbf{u}_{i2}^\top\mathbf{k}\nonumber\\
&=\frac{1}{(1+\alpha J^2(k))^2}\sum_{l=1}^{N}\rho_{k,i2}\mathbf{u}_{i2}^\top\mathbf{k}\frac{\beta\theta_{i2}(1+\alpha J^2(k))^2}{\beta\theta_{i2}+(1+\alpha J(k))^2}\nonumber\\
&=\sum_{i2=1}^{N}\rho_{k,i2}\mathbf{u}_{i2}^\top\mathbf{k}\frac{\beta\theta_{i2}}{\beta\theta_{i2}+(1+\alpha J^2(k))^2}.\nonumber
\end{align}
We observe that a nonzero $\alpha$ reduces or shrinks the contribution from the graph-frequencies corresponding to larger $J^2(k)$, in comparison with the conventional mean obtained by setting $\alpha=0$.
In the case of smooth graph signals, we have $J^2(k)=\lambda_k$ implying that the prediction mean has lower contributions from higher graph-frequencies, which in turn shows that GPG performs a noise-smoothening by making use of the graph topology. However, this does not imply that the value of $\alpha$ may be set to be arbitrarily large. A large $\alpha$ will in turn force the resulting target predictions to lie close to the smoother eigenvectors of $\mathbf{L}$, which is not desirable since it reduces the learning ability of the GP. 
 We note that since GPG-L is a special case of the GPG-K, Theorem 1 and the analysis following it apply directly also to GPG-L.

\subsection{Extension for directed graphs}
In our analysis, we have assumed the underlying graph for the target vectors to be undirected and hence, characterized by the symmetric positive semidefinite graph-Laplacian matrix $\mathbf{L}$. We now discuss how GPG may be derived for the case of directed graphs, that is, when the adjacency matrix of the graph is assymetric. In the case of directed graphs, the following metric is popular for quantifying the smoothness of the graph signals:
\begin{equation}
\mbox{MS}_g(\mathbf{y})=\|\mathbf{y}-\mathbf{A}\mathbf{y}\|_2^2,\nonumber
\end{equation}
where $\mathbf{y}$ and $\mathbf{A}$ denote the graph signal and adjacency matrix, respectively, the adjacency matrix assumed to be normalized to have maximum eigenvalue modulus equal to unity \cite{Sandry1}.
A signal $\mathbf{y}$ is smooth over a directed graph if it has a small $\mbox{MS}_g$ value. This is because $\mbox{MS}_g$ measures the difference between the signal and its one-step diffused or 'graph-shifted' version, thereby measuring the difference between the signal value at each node and its neighbours, weighted by the strength of the edge between the nodes. In the case of directed graphs, we may adopt the same approach as that of the undirected graph case with one important distinction: instead of solving \eqref{GSgen1}, we now solve the following problem for each observation:
\begin{eqnarray}
\label{GSgen2}
\mathbf{y}'_n&=&\arg\min_{\mathbf{z}}\|\mathbf{y}_n-\mathbf{z}\|_2^2+\alpha\mbox{MS}_g(\mathbf{z}),\nonumber\\
&=&\arg\min_{\mathbf{z}}\|\mathbf{y}_n-\mathbf{z}\|_2^2+\alpha\mathbf{z}^\top(\mathbf{I-A})^\top(\mathbf{I-A})\mathbf{z},\nonumber
\end{eqnarray}
Then, we have that
\begin{eqnarray}
\mathbf{y}'_n&=&\left(\mathbf{I}_M+\alpha(\mathbf{I-A})^\top(\mathbf{I-A})\right)^{-1}\mathbf{y}_n.\nonumber
\end{eqnarray}
Using a similar analysis as with the undirected graph case, we arrive at the following Gaussian process model:
\begin{equation}
\label{regprob_directed}
\tilde{\mathbf{t}}=\tilde{\mathbf{y}} + \tilde{\mathbf{e}},\nonumber
\end{equation}
where $\tilde{\mathbf{y}}$ is distributed according to the prior:
\begin{equation}
p(\tilde{\mathbf{y}})=\mathcal{N}(\tilde{\mathbf{y}}|\mathbf{0},(\mathbf{B}_d^2\otimes \mathbf{K})),\nonumber
\end{equation}
where $\mathbf{B}_d=\left(\mathbf{I}_M+\alpha(\mathbf{I-A})^\top(\mathbf{I-A})\right)^{-1}$. 
By following similar analysis as with the undirected graphs, it is possible to show that the variance of the predictive distribution reduces using GPG and that the predictive mean is smooth over the graph. 
%Similarly, by using  $\mbox{MS}_g$ instead of the quadratic form with the graph-Laplacian, we can arrive at an estimate of $\mathbf{A}$ by solving the following optimization problem:
%\begin{eqnarray}
%\label{eq: L_est_GP_A}
%\hat{\mathbf{A}}_\star&=&\arg\min_{\mathbf{A}}\left\|	\left[\mathbf{H}^{1/2}_\star\right]^\dagger-\left(\mathbf{I}_M+\alpha(\mathbf{I-A})^\top(\mathbf{I-A})\right)\right\|_F^2\nonumber\\
%&+&\omega\|\mathbf{A}\|_F^2.\nonumber
%\end{eqnarray}
In the interest of space and to avoid repetition, we do not include the analysis for directed graphs here.

\section{Experiments}
\label{GPG_experiments}
We consider application of GPG to various real-world signal examples. For the examples, we consider undirected graphs. Our interest is to compute the predictive distribution~\eqref{eq: predictive_distribution} given the noisy targets $\mathbf{T}$ and the corresponding inputs $\mathbf{X}$. Our assumption is that a target vector is smooth over an underlying graph. We use the graph-regularization with $\mathbf{G}=\mathbf{L}$.
To evaluate the prediction performance, we use the normalized-mean-square-error (NMSE) defined as follows:
\begin{equation}
\mbox{NMSE}=10\log_{10}\left(\frac{\mathcal{E}\|\mathbf{Y}-\mathbf{T}_0\|_F^2}{\mathcal{E}\|\mathbf{T}_0\|_F^2}\right),\nonumber
\end{equation}
where $\mathbf{Y}$ denotes the mean of the predictive distribution and $\mathbf{T}_0$ the true value of target matrix, that means $\mathbf{T}_0$ does not contain any noise. The noisy target matrix $\mathbf{T}$ is generated obtained by adding white Gaussian noise with precision parameter $\beta$ to $\mathbf{T}_0$.
In the case of real-world examples, we compare the performance of the following cases:
\begin{enumerate}
	\item GP with linear regression (GP-L): ${k}_{m,n}=\gamma^{-1}\pmb\phi(\mathbf{x}_m)^\top\pmb\phi(\mathbf{x}_n)$, where $\pmb\phi(\mathbf{x})=\mathbf{x}$ and $\alpha=0$,
	\item GPG with linear regression (GPG-L): ${k}_{m,n}~=~\gamma^{-1}\pmb\phi(\mathbf{x}_m)^\top\pmb\phi(\mathbf{x}_n)$ and $\alpha>0$, where $\pmb\phi(\mathbf{x})=\mathbf{x}$,
	\item GP with kernel regression (GP-K): Using radial basis function (RBF) kernel ${k}_{m,n}~=~\gamma^{-1}\displaystyle\exp\left(-\frac{\|\mathbf{x}_m-\mathbf{x}_n\|_2^2}{\sigma^2}\right)$ and $\alpha=0$, and
	\item GPG with kernel regression over graphs (GPG-K): Using RBF kernel ${k}_{m,n}~=~\gamma^{-1}\displaystyle\exp\left(-\frac{\|\mathbf{x}_m-\mathbf{x}_n\|_2^2}{\sigma^2}\right)$ and $\alpha>0$.
\end{enumerate}
We use five-fold cross-validation to obtain the values of regularization parameters $\alpha$ and $\gamma$. We set the kernel parameter $\sigma^2=\displaystyle\sum_{m,n}\|\mathbf{x}_m-\mathbf{x}_n\|_2^2$. We perform experiments under various signal-to-noise ratio (SNR) levels, and choose the precision parameter $\beta$ accordingly. 
\subsection{Prediction for fMRI in cerebellum graph}
\begin{figure*}[t]
	\centering
	$
	\begin{array}{cc}
	\subfigure[]{\includegraphics[width=2.2in]{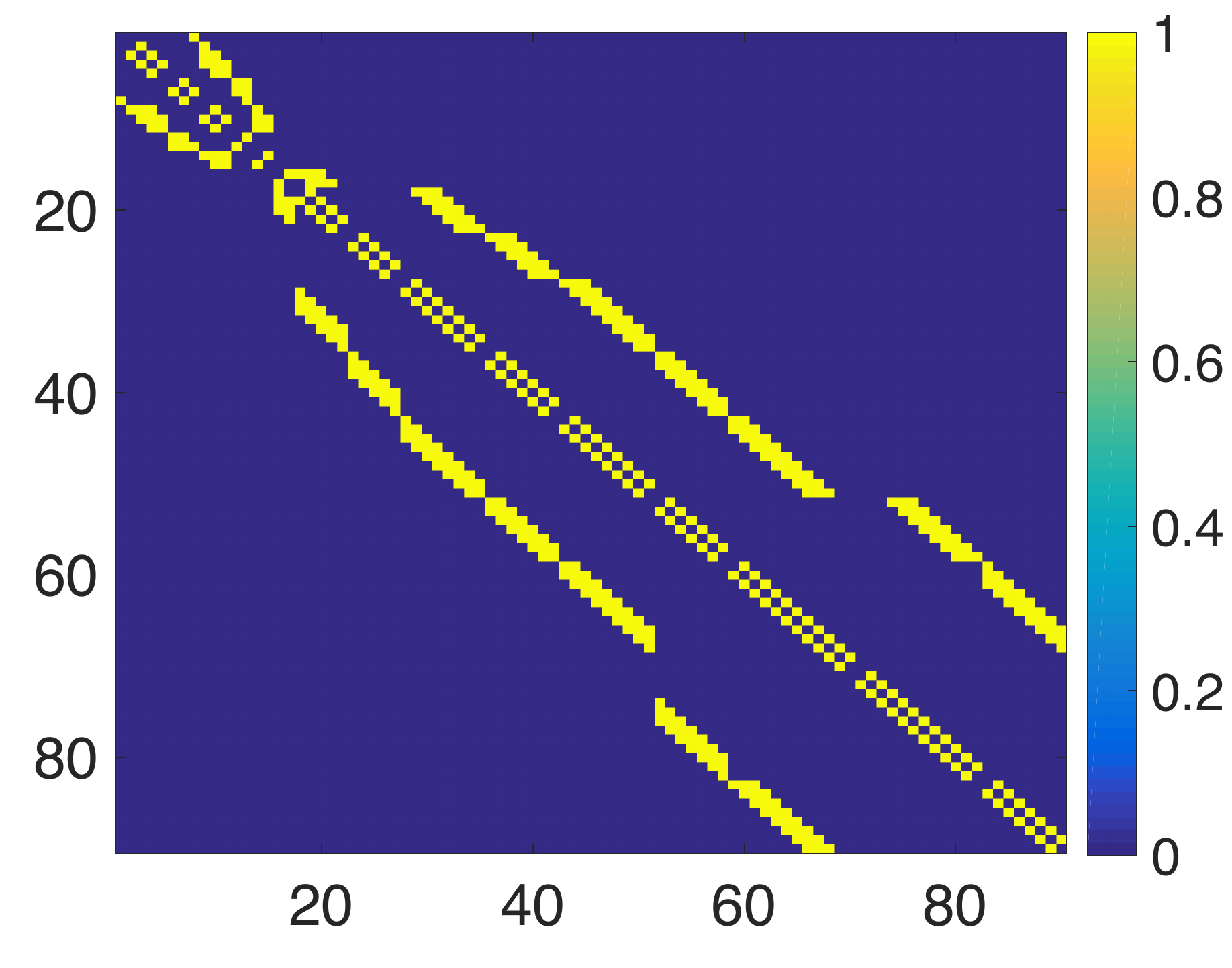}}
	\subfigure[]{\includegraphics[width=2.2in]{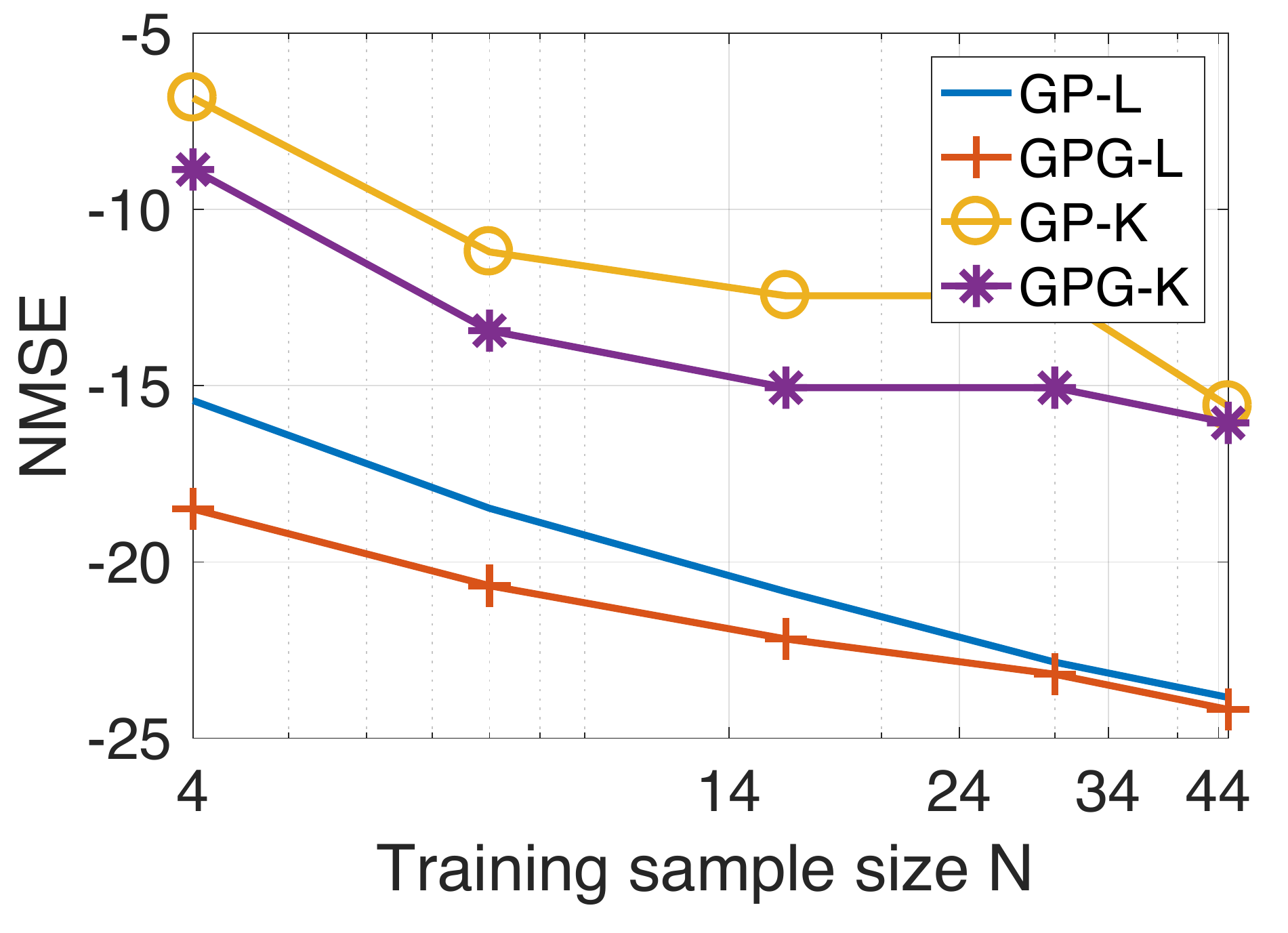}}
	\subfigure[]{\includegraphics[width=2.2in]{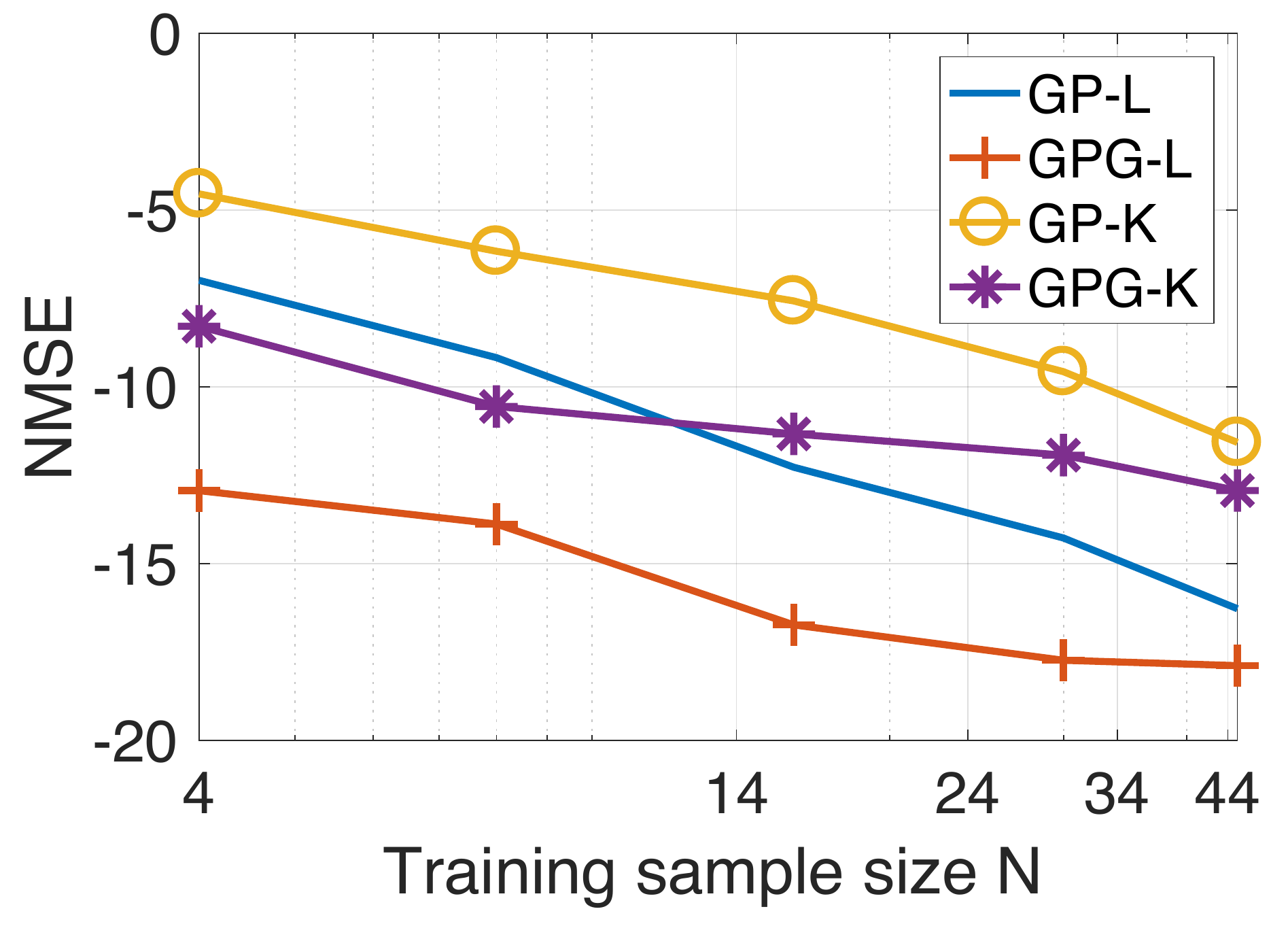}}
	\end{array}
	$
	\caption{Results for the cerebellum data  (a) Adjacency matrix, (b) NMSE for testing data as a function of training data size at SNR=$10$dB, and (c) at SNR=$0$dB.
	}
	\label{fig:GPG_Cere}
\end{figure*}

We first consider the functional magnetic resonance imaging (fMRI) data obtained for the cerebellum region of brain \cite{Behjat_1}\footnote{The data is available publicly at https://openfmri.org/dataset/ds000102.}. The original graph consists of 4465 nodes corresponding to different voxels of the cerebellum region. The voxels are mapped anatomically following the atlas template \cite{Behjat_cerebellum,cerebellum_atlas}. We refer to \cite{Behjat_1} for details of graph construction and associated signal extraction. We consider a subset of the first 100 vertices in our analysis. Our goal is to use the first ten vertices as input $\mathbf{x}\in\mathbb{R}^{10}$ to make predictions for remaining 90 vertices, which forms the output $\mathbf{t}\in\mathbb{R}^{90}$. Thus, the target signals lie over a graph of dimension $M=90$. The corresponding adjacency matrix is shown in Figure \ref{fig:GPG_Cere}(a). We have a total of 295 graph signals corresponding to different measurements from a single subject. We use a portion of the signals for training and the remaining for testing. We construct noisy training targets by adding white Gaussian noise at SNR-levels of 10 dB and 0 dB. The NMSE of the prediction mean for testing data, averaged over 100 different random choices of training and testing sets is shown in \ref{fig:GPG_Cere}(b) and (c); this is Monte Carlo simulation to check robustness. We observe that for both linear and kernel regression cases, GPG outperforms GP by a significant margin, particularly at small training data sizes as expected. The trend is also similar when larger subsets of nodes from the full set are considered. The results are not reported here for brevity and to avoid repetition.
\subsection{Prediction for temperature data}
\begin{figure*}
	\centering
	$
	\begin{array}{cc}
	\subfigure[]{\includegraphics[width=2.2in]{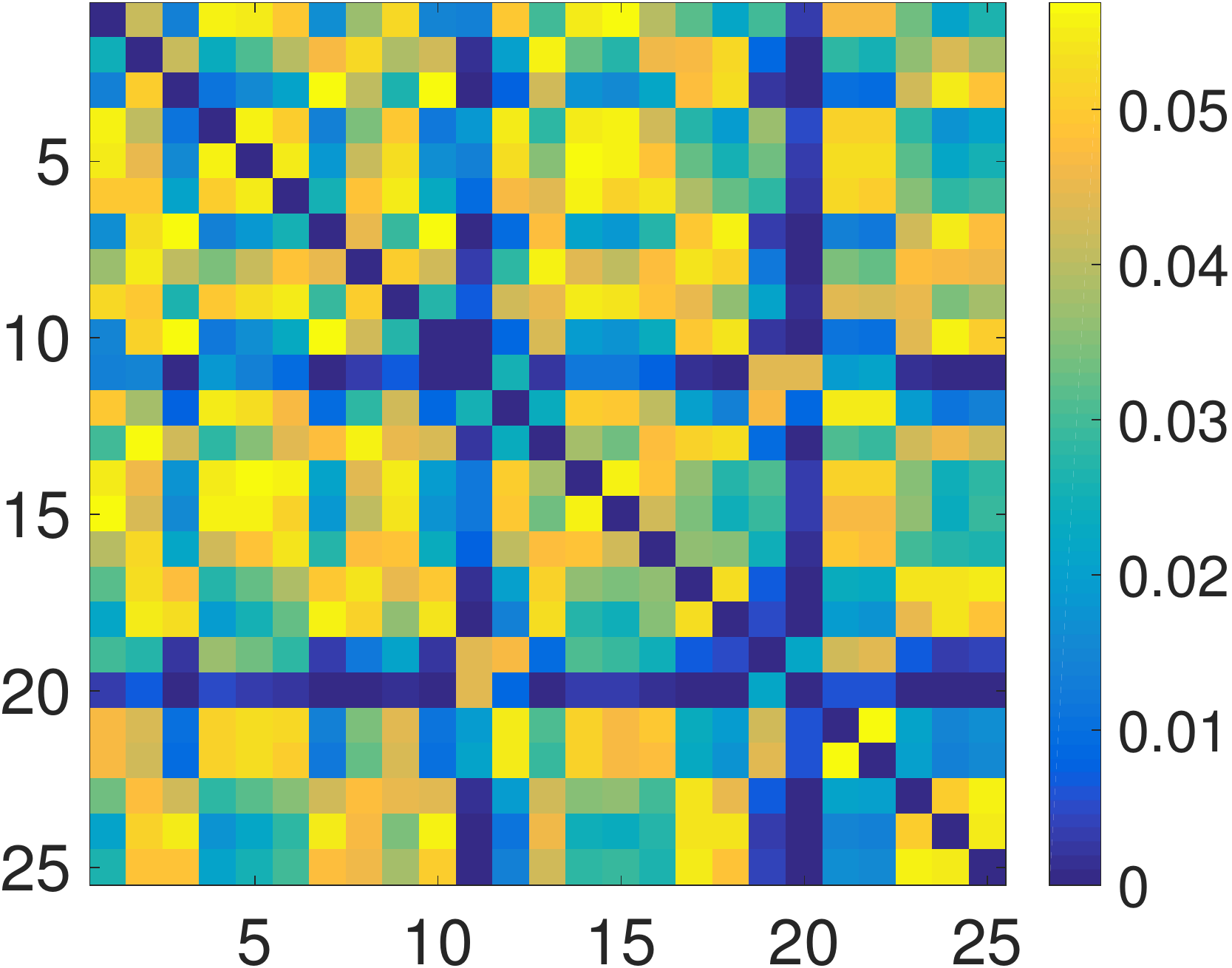}}
	\subfigure[]{\includegraphics[width=2.2in]{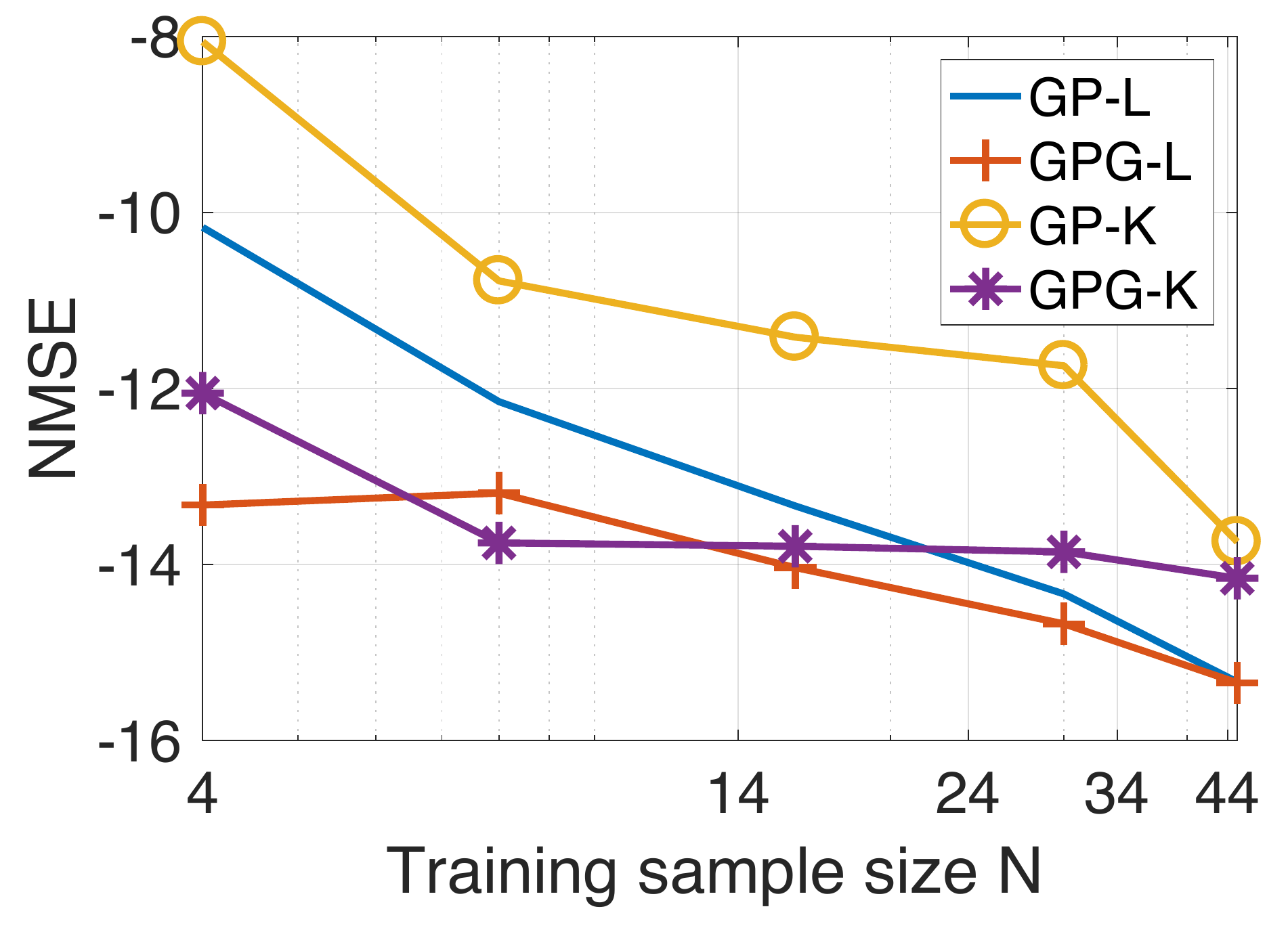}}
	\subfigure[]{\includegraphics[width=2.2in]{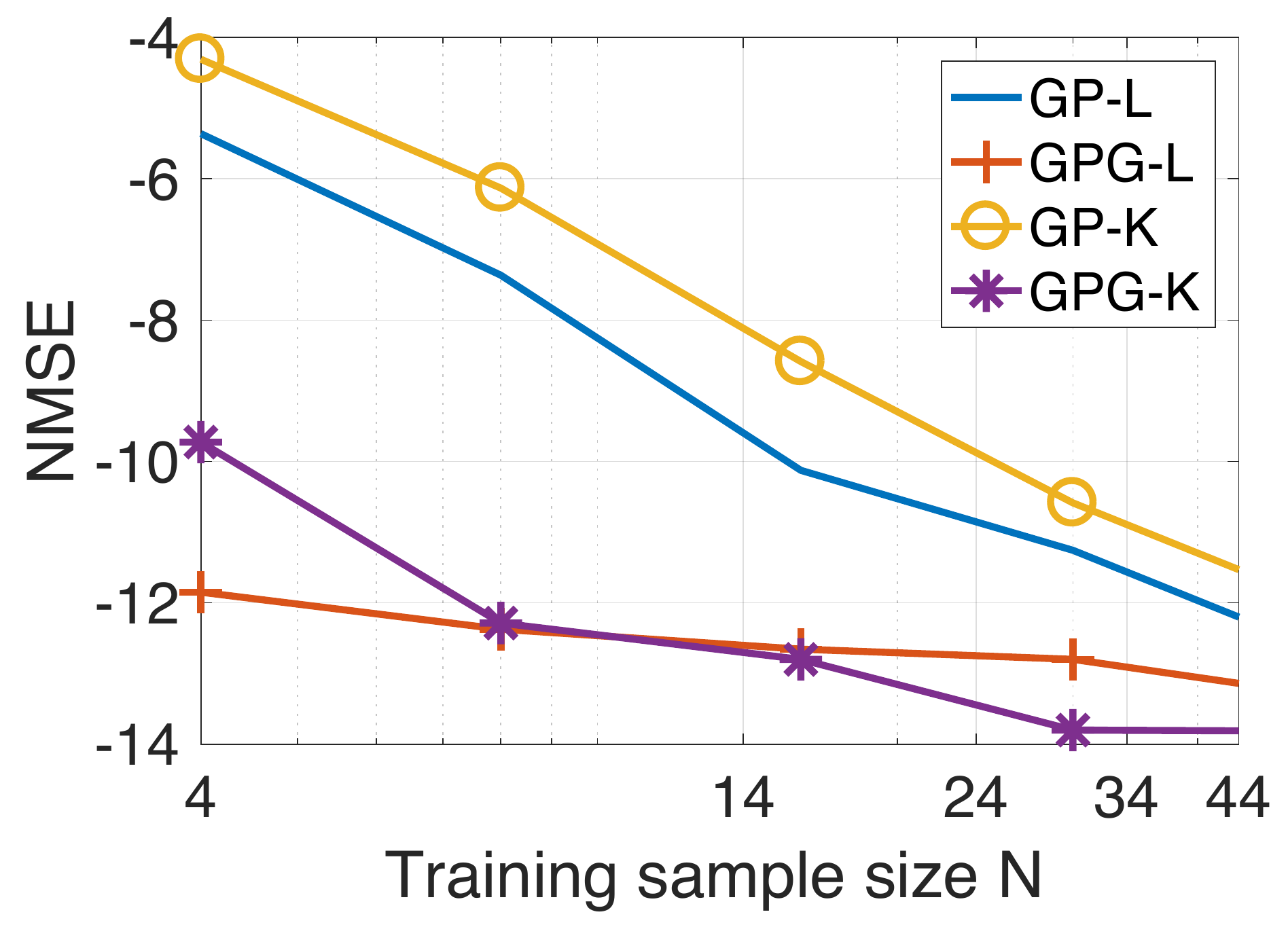}}
	\end{array}
	$
	\caption{Results for the temperature data (a) Adjacency matrix, (b) NMSE for testing data as a function of training data size at SNR=$5$dB, and (c) at SNR=$0$dB.
	}
	\label{fig:GPG_Temp}
\end{figure*}

We next apply GPG on temperature measurements from the 45 most populated cities in Sweden for a period of three months from October to December 2017. The data is available publicly from the Swedish Meteorological and Hydrological Institute \cite{SMHI}. Our goal is to perform one-day temperature prediction: given the temperature data for a particular day $\mathbf{x}_n$, we predict the temperature for the next day $\mathbf{t}_n$. We have 90 input-target data pairs in total, of which one half is used for training and the rest for testing. We consider the geodesic graph in our analysis with the $(i,j)$th entry of the graph adjacency matrix $\mathbf{A}$ is given by
\begin{equation}
\mathbf{A}(i,j)=\exp{\left(-\frac{d_{ij}^2}{\sum_{i,j}d_{ij}^2}\right)},\nonumber
\end{equation}
where $d_{ij}$ denotes the geodesic distance between the $i$th and $j$th cities. In order to remove self loops, the diagonal of $\mathbf{A}$ is set to zero. We generate noisy training data by adding zero-mean white Gaussian noise at SNR of $5$ dB and $0$ dB to the true temperature measurements. In Figure \ref{fig:GPG_Temp}, we show the NMSE obtained for testing data by averaging over 100 different random partitioning of the total dataset into training and testing datasets. We observe that GPG outperforms GP for both linear and kernel regression cases.
\subsection{Prediction for flow-cytometry data}
\begin{figure*}
	\centering
	$
	\begin{array}{cc}
	\subfigure[]{\includegraphics[width=2.in]{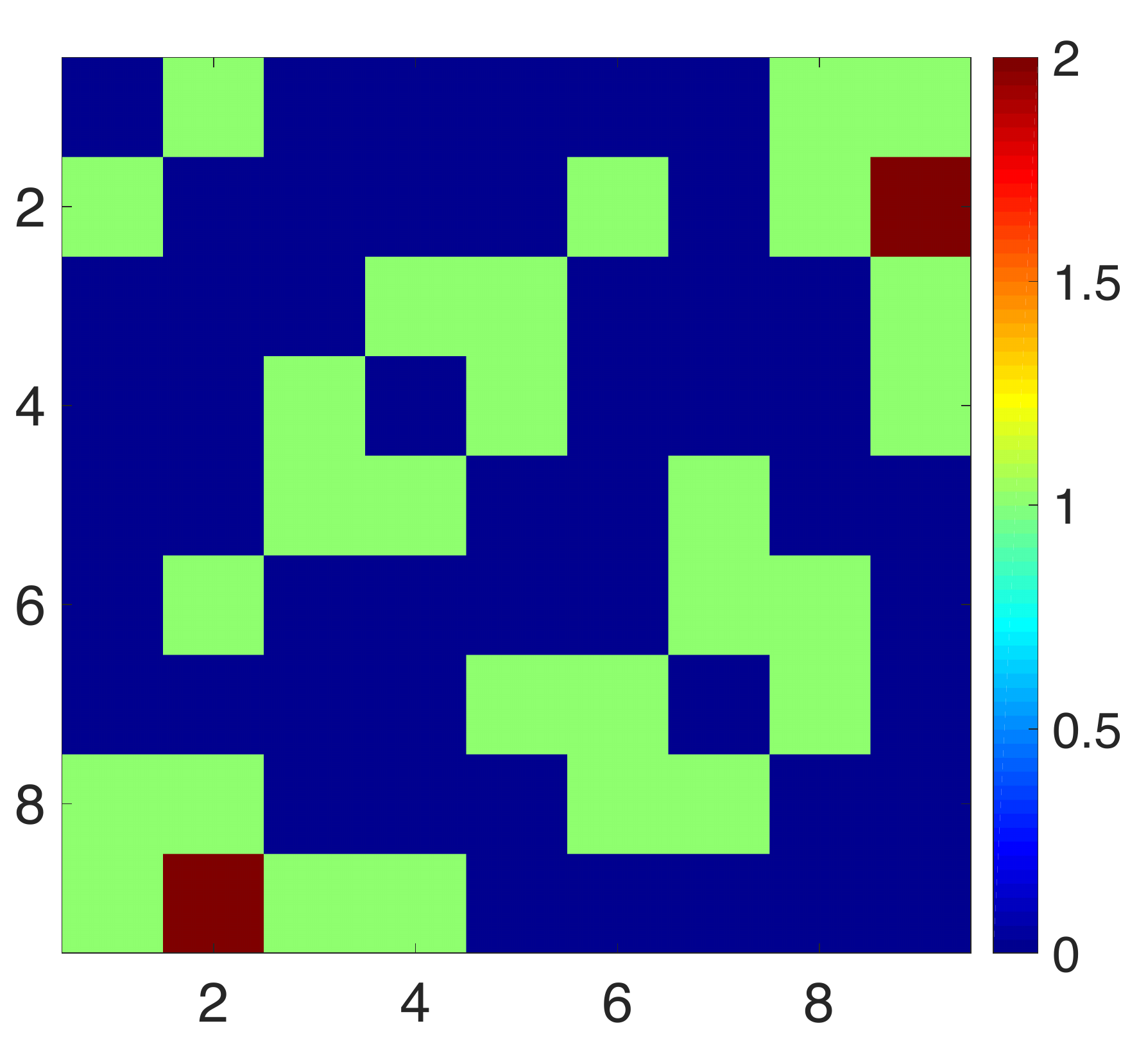}}
	\subfigure[]{\includegraphics[width=2.2in]{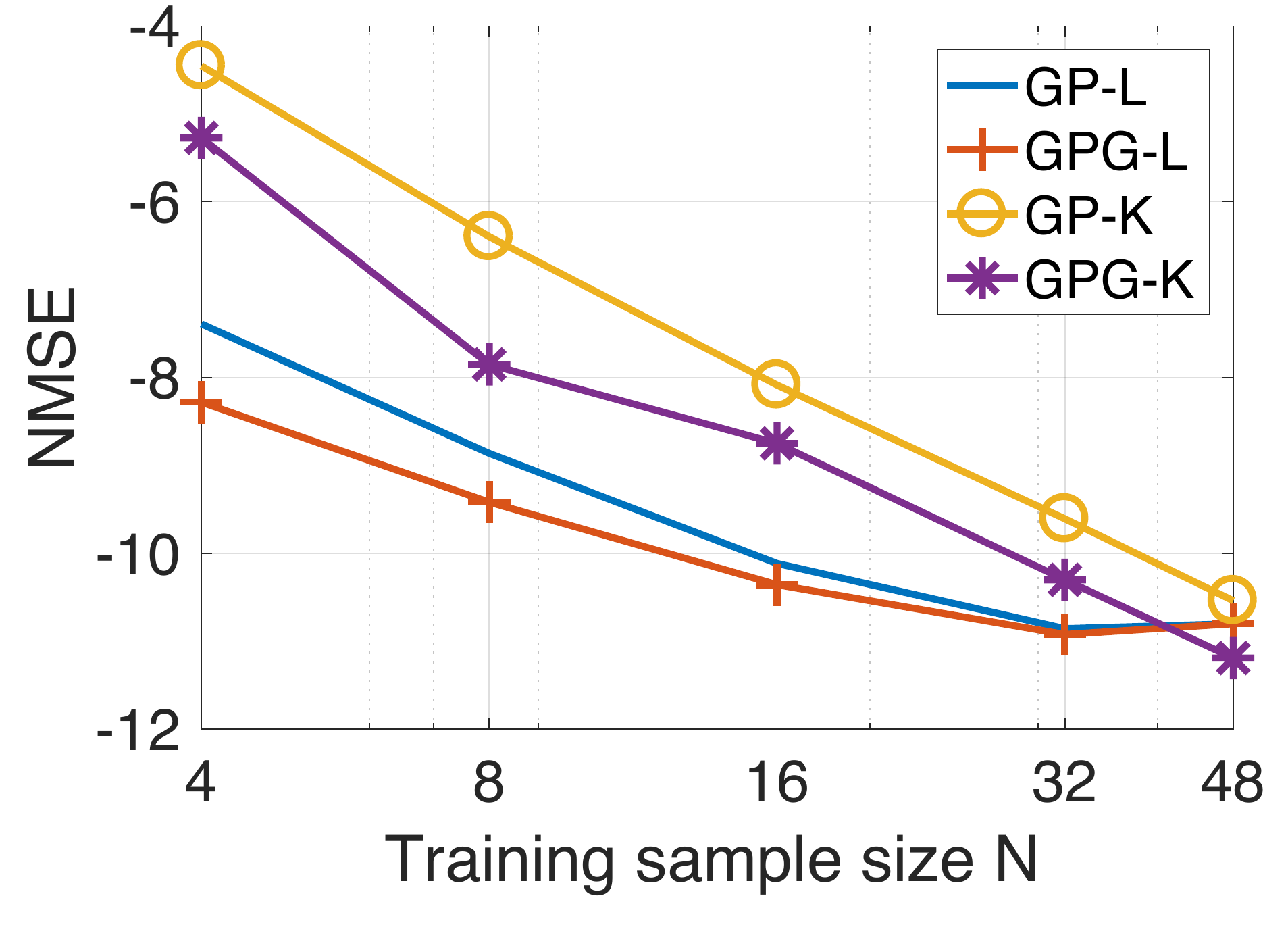}}
	\subfigure[]{\includegraphics[width=2.2in]{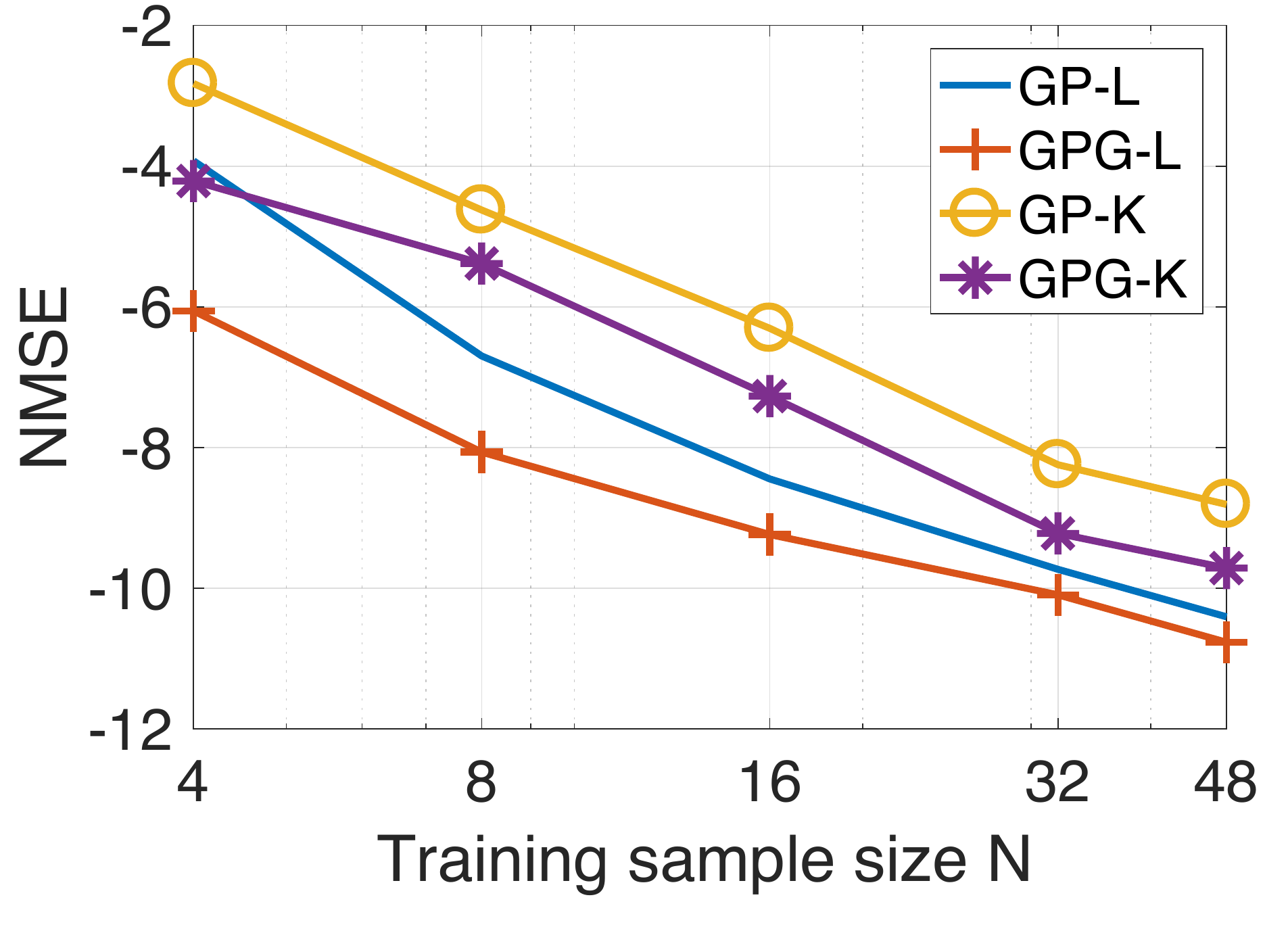}}
	\end{array}
	$
	\caption{Results for flow-cytometry data (a) Adjacency matrix, (b) NMSE for testing data as a function of training data size at SNR=$5$dB, and (c) at SNR=$0$dB. }
	\label{fig:GPG_Cell}
\end{figure*}

\begin{figure*}
	\centering
	$
	\begin{array}{cc}
	\subfigure[]{\includegraphics[width=2.2in]{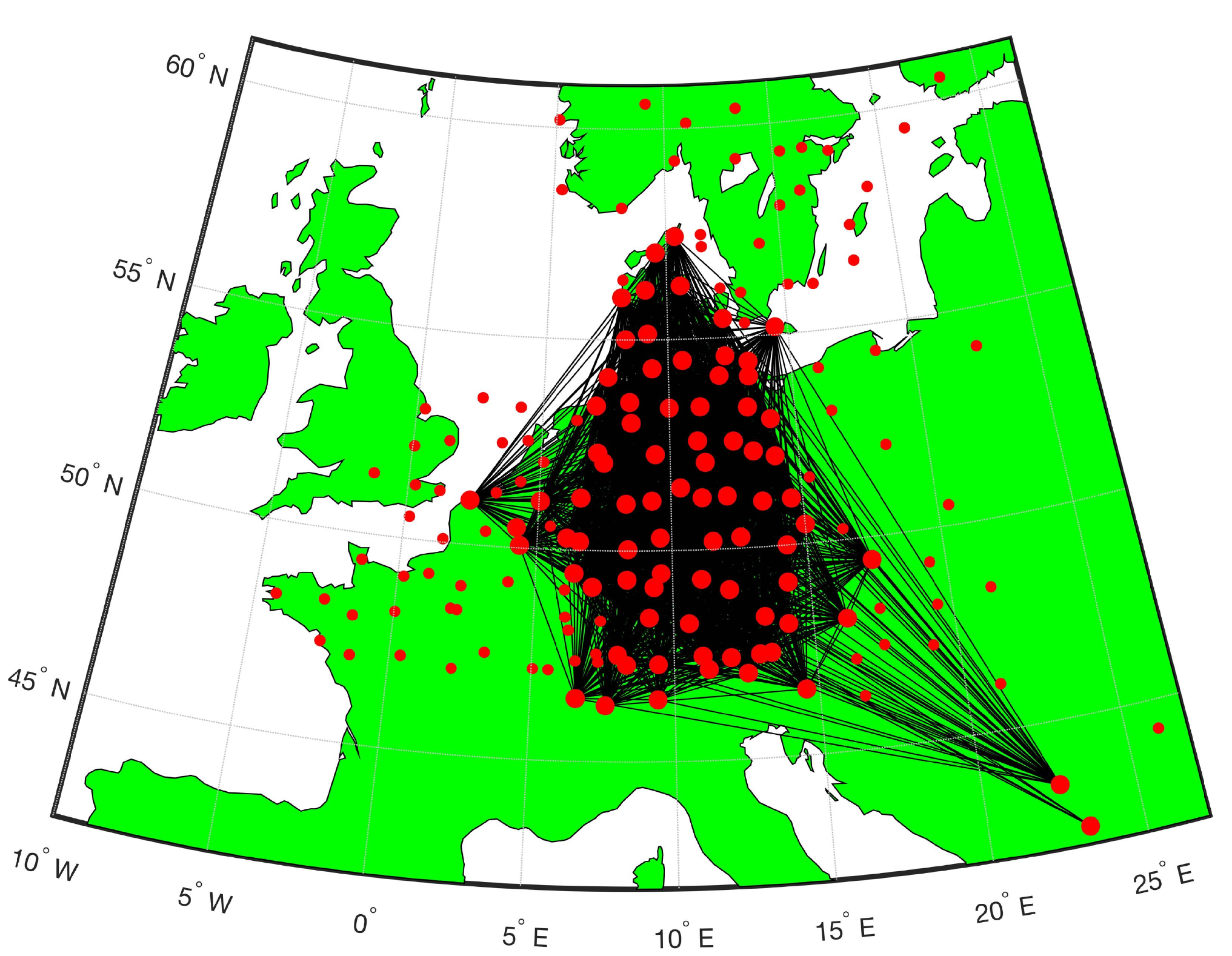}}
	\subfigure[]{\includegraphics[width=2.2in]{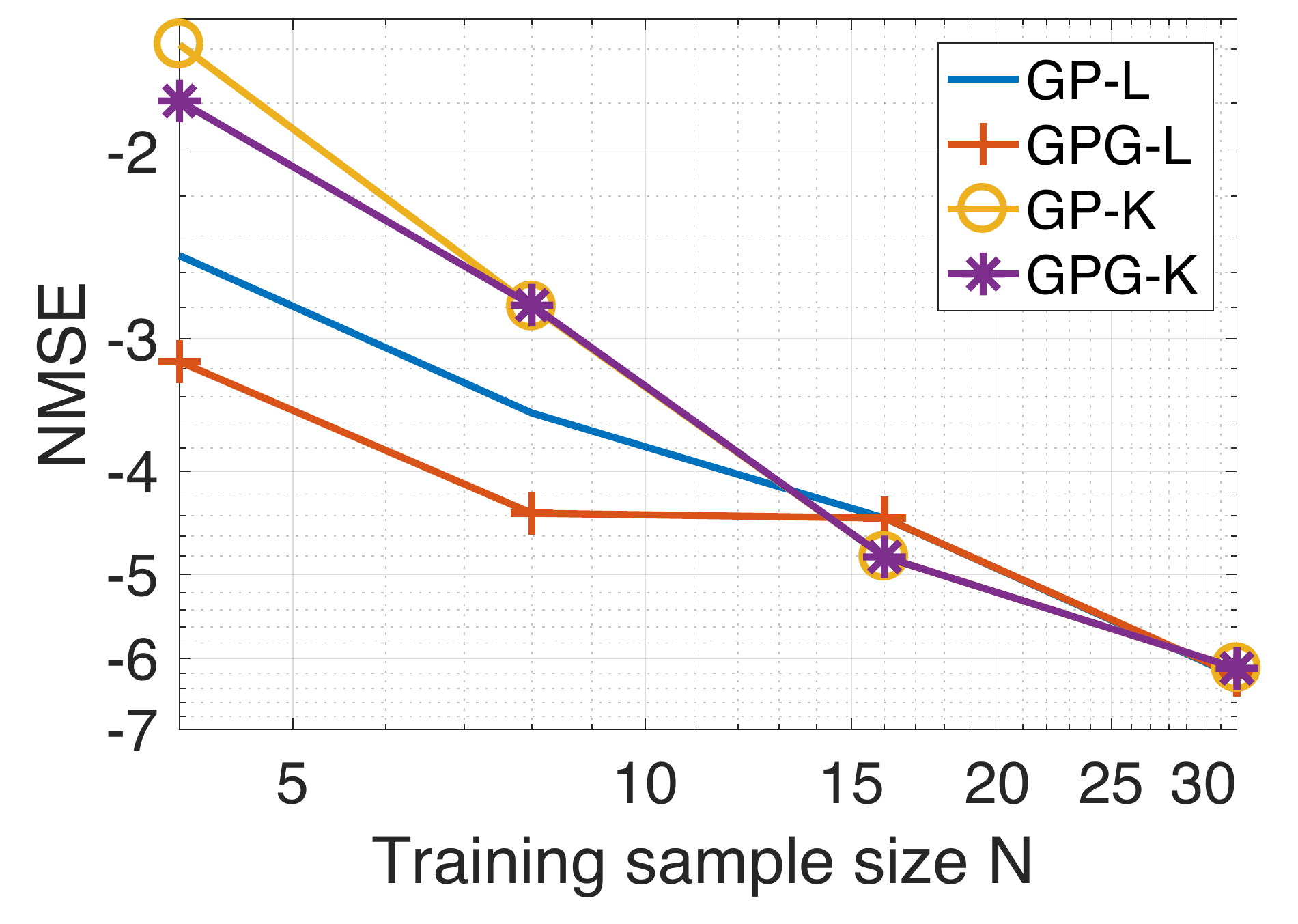}}
	\subfigure[]{\includegraphics[width=2.2in]{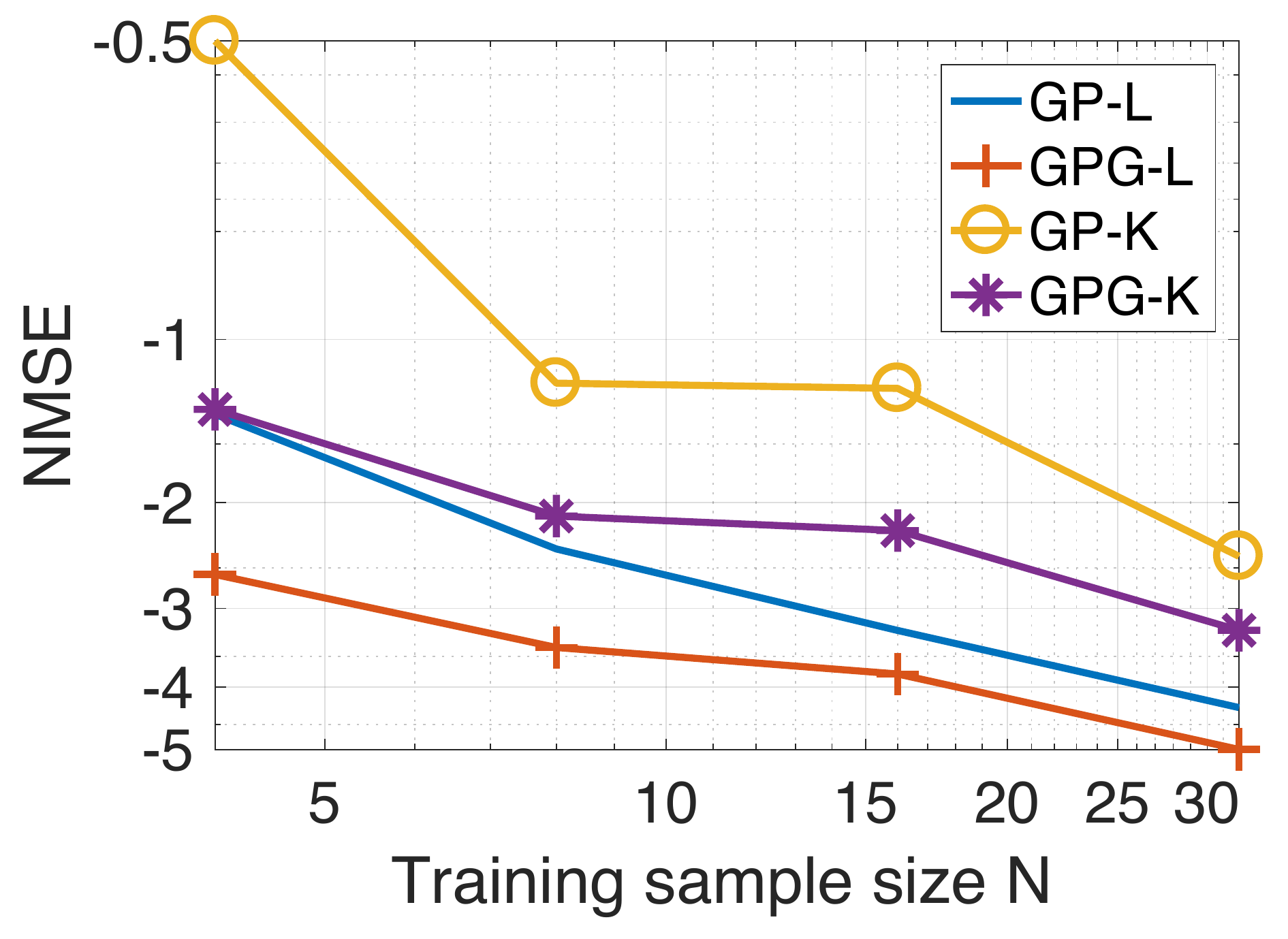}}
	\end{array}
	$
	\caption{Results for ETEX data (a) Schematic showing the ground stations for the experiment, (b) NMSE for testing data as a function of training data size at SNR=$5$ dB, and (c) at SNR=$0$ dB. }
	\label{fig:GPG_ETEX}
\end{figure*}

\begin{table*}
	\label{proteinnames}
	\centering
	\begin{tabular}{|c|c|c|c|c|c|c|c|c|c|c|c|c|c|c}
		\hline
		Number& 1 &2&3&4&5&6&7&8&9&10&11\\
		\hline
		Name&praf&	pmek&	plcg&	PIP2&	PIP3&	p44/42	&pjnk
		&pakts473&	PKA&	PKC&	P38\\
		\hline
	\end{tabular}
	\caption{Names of different proteins that represent the nodes of the graph considered in Section IV-D.}
\end{table*}
We now consider the application of GPG to flow-cytometry data considered by Sachs et al. \cite{Sachs} which consists of response or signalling level of 11 proteins in different experiment cells. Since the protein signal values have a large dynamic range and are all positive values, we perform experiments on signals obtained by taking a logarithm with the base of 10 for reducing the dynamic range. We use the first 1000 measurements in our analysis. We use the symmetricized version of the directed unweighted acyclic graph proposed by Sachs et al. \cite{Sachs}. Among the 11 proteins, we choose proteins 10 and 11 arbitrarily as input to make predictions for the remaining 9 proteins which forms the target vector $\mathbf{y}_n\in\mathbb{R}^7$ lying on a graph of $M=7$ nodes. We perform the experiment 100 times in Monte Carlo simulation, where we random divide the total dataset into training and testing datasets. The average NMSE for testing datasets is shown in Figure \ref{fig:GPG_Cell} at SNR levels of $5$ dB and $0$ dB. We once again observe the same trend that GPG outperforms GP for both linear and kernel cases at low sample sizes corrupted with noise.
{\subsection{Prediction for atmospheric tracer diffusion data}
	Our next experiment is on the atmospheric tracer diffusion measurements obtained from the European Tracer Experiment (ETEX) which tracked the concentration of perfluorocarbon tracers released into the atmosphere starting from a fixed location (Rennes, France)\cite{ETEX}. The observations were collected from two experiments over a span of 72 hours at 168 ground stations over Europe, giving two sets of 30 measurements, in total 60 measurements. Our goal is to predict the tracer concentrations on one half (84) of the total ground stations, using the concentrations at the remaining locations. The target signal $\mathbf{t}_n$ is then a graph signal over a graph of $M=84$ nodes where $n$ denotes the measurement index. The corresponding input vector $\mathbf{x}_n$ is also of length 84. We illustrate the ground station locations in a schematic in Figure \ref{fig:GPG_ETEX} (a). The output nodes which correspond to the the target are shown in red markers with corresponding edges, whereas the rest of the markers denote the input. We simulate noisy training by adding white Gaussian noise at different SNR levels to the training data. We consider a geodesic distance based graph. The graph is constructed in the same manner like the graph used in the temperature data experiment previously. We randomly divide the total dataset of 60 samples equally into training and test datasets. We compute the NMSE for the different approaches by averaging over 100 different randomly drawn training subsets of size $N$ from the full training set of size $N_{ts}=30$. We plot the NMSE as a function of $N$ in Figures \ref{fig:GPG_ETEX}(b)-(c) at SNR levels of $5$dB and $0$dB. We observe that graph structure enhances the prediction performance signficantly under noisy and low sample size conditions.

}

	\section{Reproducible research}
In the spirit of reproducible research, all the codes relevant to the experiments in this article are made available at https://www.researchgate.net/profile/Arun\textunderscore Venkitaraman and https://www.kth.se/ise/research/reproducibleresearch-1.433797.

\section{Conclusions}
We developed Gaussian processes for signals over graphs by employing a graph-Laplacian based regularization. The Gaussian process over graphs was shown to be a consistent generalization of the conventional Gaussian process, and that it provably results in a reduction of uncertainty in the output prediction. This in turn implies that the Gaussian process over graph is a better model for target vectors lying over a graph in comparison to the conventional Gaussian process. This observation is important in cases when the available training data is limited in both quantity and quality.
Our expectation and motivation was that incorporating the graph structural information would help the Gaussian process make better predictions, particularly in absence sufficient and reliable training data. The experimental results with the real-world graph signals illustrated that this is indeed the case. 
%We also observed that considerable information about the network may be extracted as a by-product of the Gausian process over graph in a maximum-likelihood setting with even moderate number of training observations in comparison with the signal dimensions.
	
				\bibliographystyle{IEEEtran}
				\bibliography{refs,refs_2}
	\end{document}